\documentclass[11pt]{article}

\usepackage[preprint]{acl}

\usepackage{times}
\usepackage{latexsym}
\usepackage{amsmath}
\usepackage{amssymb}
\usepackage{amsfonts}
\usepackage{mathtools}

\usepackage[T1]{fontenc}
\usepackage[utf8]{inputenc}

\usepackage{microtype}
\usepackage{inconsolata}

\usepackage{graphicx}
\usepackage{xspace}
\usepackage{makecell}
\usepackage{array}
\usepackage{colortbl}
\usepackage{multirow}
\usepackage{subfig}
\usepackage{tabularx}
\usepackage[dvipsnames]{xcolor}
\usepackage{color}
\usepackage{pifont}
\usepackage{enumitem}
\usepackage{diagbox}
\usepackage{listings}
\usepackage[flushleft]{threeparttable}
\usepackage{booktabs}
\usepackage{wrapfig}
\usepackage{marvosym}
\usepackage{tikz}
\usepackage[listings,most]{tcolorbox}

\newcommand{\icon}{\raisebox{-4.1pt}{\includegraphics[width=1.3em]{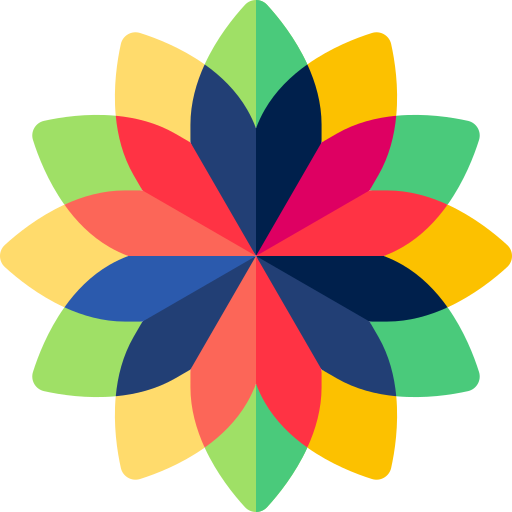}}\xspace}

\definecolor{fbApp}{HTML}{ffe4e3}
\definecolor{mydarkblue}{rgb}{0,0.3,0.9}

\newcommand{\rowcr}{\rowcolor{fbApp}}

\newcommand{\rowcb}{\rowcolor{CadetBlue!10}}
\newcommand{\rowcg}{\rowcolor{gray!10}}

\newcommand{\first}[1]{\textcolor{red}{\textbf{#1}}}
\newcommand{\second}[1]{\textcolor{blue}{\underline{#1}}}

\title{\icon TimeART: Towards Agentic Time Series Reasoning via Tool-Augmentation}

\author{
  Xingjian Wu, Junkai Lu, Zhengyu Li, Xiangfei Qiu, Jilin Hu, \\ \textbf{Chenjuan Guo}, \textbf{Christian S. Jensen}, \textbf{Bin Yang}\textsuperscript{\Letter}
\\
 East China Normal University
\\
   \{xjwu, jklu, lizhengyu, xfqiu\}@stu.ecnu.edu.cn, \\ 
   \{jlhu, cjguo, byang\}@dase.ecnu.edu.cn
}

\begin{document}
\maketitle

\begin{abstract}
Time series data widely exist in real-world cyber-physical systems. Though analyzing and interpreting them contributes to significant values, e.g, disaster prediction and financial risk control, current workflows mainly rely on human data scientists, which requires significant labor costs and lacks automation. To tackle this, we introduce TimeART, a framework fusing the analytical capability of strong out-of-the-box tools and the reasoning capability of Large Language Models (LLMs), which serves as a tool-augmented reasoning system for Time Series Question Answering (TSQA). To teach the LLM-based Time Series Reasoning Models (TSRMs) strategic tool-use, we also collect a \textit{100k} expert trajectory corpus called TimeToolBench. To enhance TSRMs' generalization capability, we then devise a four-stage training strategy, which boosts TSRMs through learning from their own early experiences and self-reflections. Experimentally, we train an 8B TSRM on TimeToolBench and equip it with the TimeART framework, and it achieves state-of-the-art performance on multiple TSQA tasks, which demonstrates the effectiveness of tool-augmented training for TSQA.
\end{abstract}

\section{Introduction}
Time series data widely exist in real-world cyber-physical systems from transportation, finance, healthcare, weather, and other domains~\citep{qiu2024tfb,qiu2025tab,li2025TSFM-Bench}. Analyzing such time series data can lead to deep insights into the inherent dynamics, and boost decision-making in downstream tasks~\citep{qiu2025duet,wu2025k2vae,wu2024catch,wu2025srsnet}. In conventional pipelines, data scientists play a vital role in process orchestration and model design. Though facilitating the promotion of time series analysis, they heavily rely on expert prior knowledge and lack automation. To explore a fully-agentic paradigm which can work as smartly as human experts, recent studies focus on unleashing the potential of LLMs in time series analysis through constructing reasoning corpus, covering analytical and reasoning tasks~\citep{TimeMQA,cai2024timeseriesexam}, and then training Time Series Reasoning Models (TSRMs) on them~\citep{xie2024chatts,sui2025training}. However, there still remain limitations in TSRMs, making them struggle to replace data scientists.

\begin{figure}[t]
    \centering
    \includegraphics[width=\linewidth]{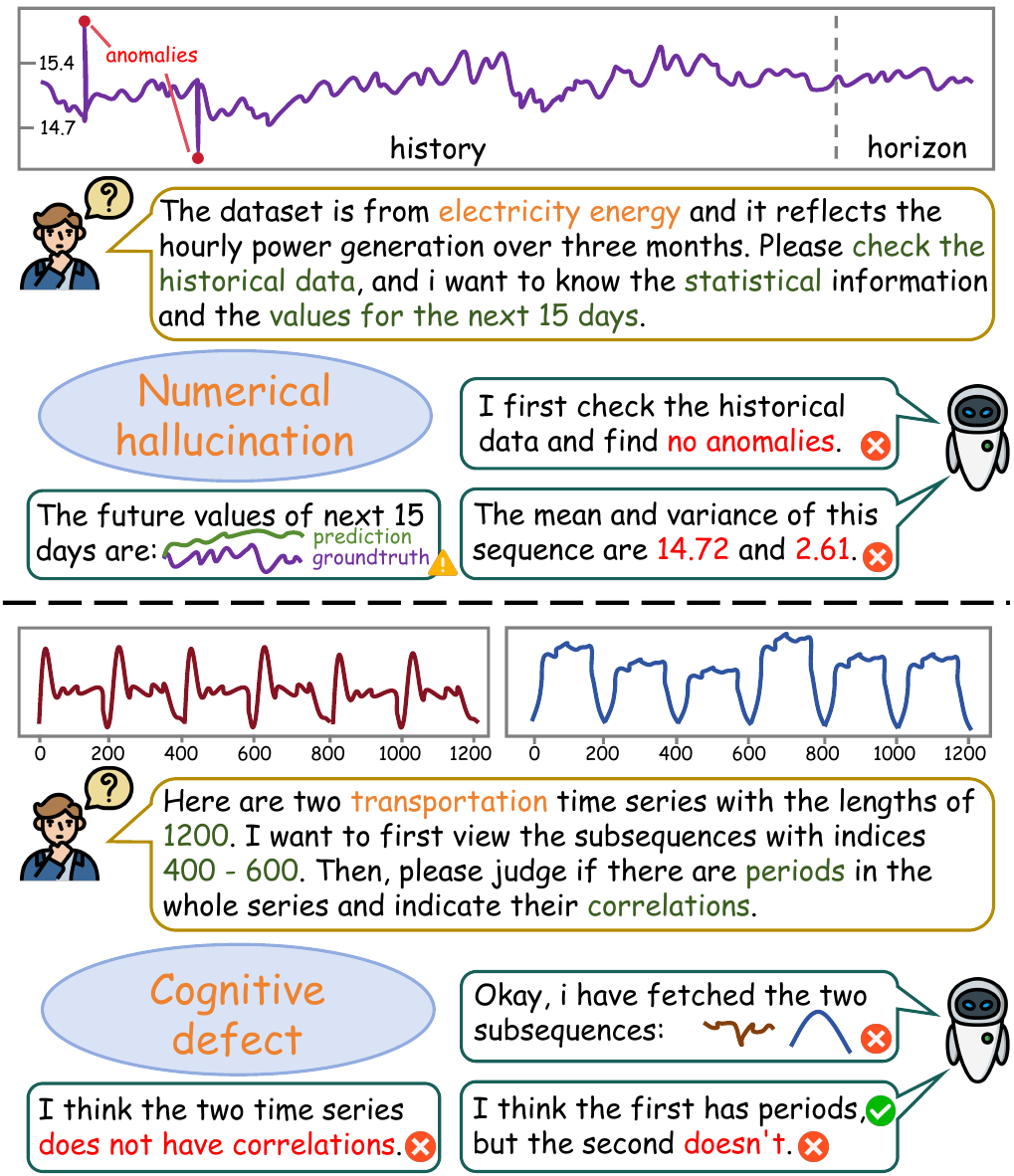}
    \caption{Common dilemmas in time series reasoning, i.e., \textit{numerical hallucination} and \textit{cognitive defect}.}
\label{fig: intro1}
\end{figure}

Current studies on TSRMs~\citep{xie2024chatts,TimeMQA} aim to strengthen and unleash the inherent capabilities of LLMs in time series reasoning tasks, which requires LLMs to fully understand the internal connections of structured numerical values. Due to the discrete tokenization of LLMs, they often struggle to handle such numerical tasks and prove suboptimal in analytical tasks like forecasting~\citep{tan2024language} and anomaly detection~\citep{zhou2024can}. In reasoning tasks, they also face two main problems in processing time series data--see Figure~\ref{fig: intro1}: 1) \textit{numerical hallucination}. TSRMs cannot accurately understand and calculate statistical characteristics of time series, nor make reliable forecasts and anomaly detections; 2) \textit{cognitive defect}. When the input time series is long, and the questions are flexible and complex, TSRMs struggle to provide correct analytics. From our perspective, \textit{simulating data scientists with TSRMs but not providing strong analytical tools they often use} does hinder the capabilities.

To overcome the above-mentioned problems, we propose \textbf{TimeART}, a \textbf{\underline{Time}} series \textbf{\underline{A}}gentic \textbf{\underline{R}}easoning framework with \textbf{\underline{T}}ool-augmentation. It integrates \textit{21} strong out-of-the-box tools for time series analysis, which include statistical methods to lightweight time series foundation models~\citep{wang2025lightgts,shentu2024towards}, allowing TSRMs to autonomously, robustly, and efficiently use tools in the reasoning process. Furthermore, in order to strengthen the tool invocation capability of TSRMs, we propose \textbf{TimeToolBench}, a comprehensive corpus for agentic time series reasoning, containing \textit{100k} ReAct-style~\citep{yao2022react} tool-use trajectories from multiple real-world domains, collected with GPT-4o and validated through a strict scheme. This provides a general solution to finetune any open-source LLM with this corpus to strong TSRM supporting tool-use.

\begin{figure}[t]
    \centering
\includegraphics[width=\linewidth]{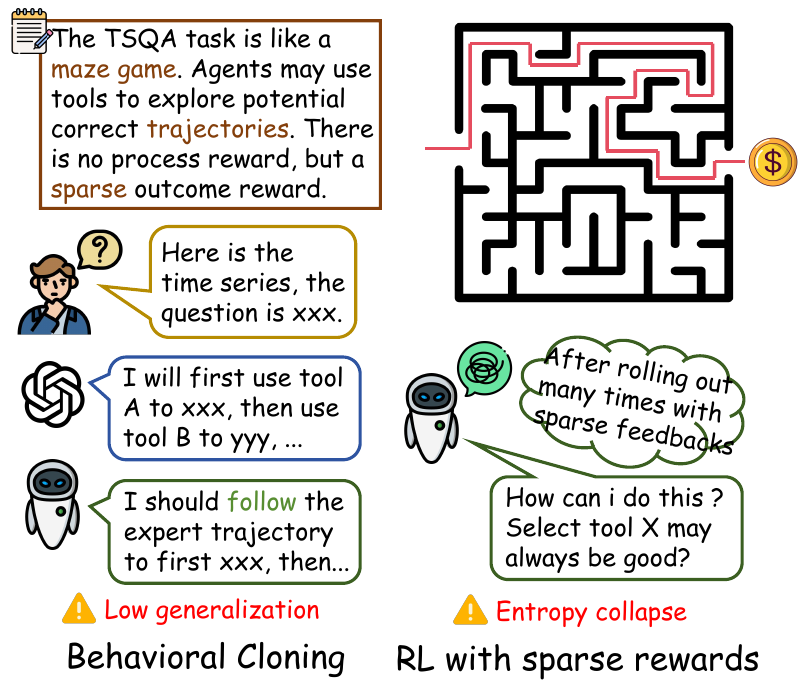}
    \caption{Common dilemmas in behaviour cloning and reinforcement learning, causing \textit{low generalization} and \textit{entropy collapse}.}
\label{fig: intro2}
\end{figure}

On the other hand, the training paradigm of TSRMs with tool-use also matters. Current training paradigm from other domains, e.g. math reasoning agent~\citep{forootani2025survey}, GUI agent~\citep{zhang2024large}, and web agent~\citep{ning2025survey}, mainly adopts first behaviour cloning then reinforcement learning. In these domains, the extent of behaviour cloning is widely studied and the process reward of reinforcement learning is clearly-defined. However, this paradigm faces some challenges in time series reasoning--see Figure~\ref{fig: intro2}: 1) \textit{low generalization.} The balance of behaviour cloning is hard to control, often causing TSRMs to simply imitate the tool-use trajectories of experts. 2) \textit{entropy collapse.} The reward of reinforcement learning in time series reasoning tasks is sparse, lacking the process reward and only possessing the outcome reward. This makes TSRMs hard to understand the environment (world modeling) and tend to output conservative decisions.

The underlying principle of agentic reinforcement learning~\citep{zhang2509landscape} is to implicitly align the preferences of the reward functions with the model, allowing it to model the endogenous world for preference-based decision-making. However, it is limited by the design of reward functions and quality of rollouts, especially when the reward is sparse. To tackle this, we draw inspiration from experience learning~\citep{zhang2025agent,fang2025play2prompt,tang2025chemagent}, and first devise multiple stages to train the TSRMs for agentic reasoning, which allows them to learn from early exploration experience and self-reflection. The training strategy boosts perceiving the action space and the external environment, learning the strategic tool-use, and understanding the underlying principles. It functionally replaces conventional (SFT + RL) process and is easier to train, which are experimentally demonstrated in Section~\ref{sec: model analysis}, thus enhancing the generalization and robustness in endowing TSRMs with the capability to invoke proper tools according to scenarios. Our contributions lie in:

\begin{itemize}[left=0.1cm,topsep=2pt,itemsep=2pt]
\item We introduce \textbf{TimeART}, an Agentic Time Series Reasoning framework supporting tool-use. It integrates \textit{21} strong out-of-the-box analytical tools, allowing TSRMs to autonomously, robustly, and efficiently invoke them during reasoning. TSRMs can be further strengthened through training on the proposed \textbf{TimeToolBench}, a corpus containing \textit{100k} high-quality tool-use trajectories.

\item We devise a novel training strategy to mitigate the dilemmas in conventional paradigms. It decomposes the underlying principles of agentic reinforcement learning into several parts, enabling the TSRMs to better perceive the external environment and the action space, thus preserving the generalization capability. 

\item Experimentally, we train an 8B TSRM on \textbf{TimeToolBench} through our proposed training strategy, and then equip it with the \textbf{TimeART} framework, which demonstrates state-of-the-art performance. The code is available at \href{https://anonymous.4open.science/r/TimeART-D004}{https://anonymous.4open.science/r/TimeART-D004}.

\end{itemize}

\section{Methodology}

\subsection{TimeART}
\label{sec: timeart}
We make ReAct-style definitions of agentic time series reasoning for clear clarification, where the reasoning process is strictly organized as five states: Query (Q), Thought (T), Action (A), Observation (O) and Final Answer (F). Among them, Thought, Action and Observation are intermediate states, which can be repeated multiple times until the Final Answer is given. The Query is given as the task descriptions and input time series, then LLM needs to think step-by-step to give the Final Answer. During the reasoning process, the steps are refined into Thought, Action, and Observation, where Thought and Observation are textual responses, and the Action is the tool-use decision with parameters equipped. In TimeART, the Action space consists of \textit{21} out-of-the-box analytical tools, and the inputs and parameters of tools are autonomously organized by TSRMs. Given a TSRM $p_\theta$, the agentic reasoning process can be described as a trajectory $\mathcal{T} := (\mathrm{Q}, (\mathrm{T}_1, \mathrm{A}_1, \mathrm{O}_1),\cdots,(\mathrm{T}_K, \mathrm{A}_K, \mathrm{O}_K), \mathrm{F})$, where each state $\mathrm{S}_i \in \mathcal{T}$ is sampled from the conditional probability distribution modeled by TSRM:
\begin{gather}
    \mathrm{S}_i \sim p_\theta (\mathrm{S}_i|\mathrm{S}_{1:i-1}, \mathrm{E}),
\end{gather}
where $\mathrm{S}_i$ can be Query (Q), Thought (T), Action (A), Observation (O) or Final Answer (F). $\mathrm{E}$ are the external constraints, such as format prompts, and tool descriptions, 
enabling TimeART to robustly invoke the tools and output the reasoning trajectory step-by-step in ReAct-style. 

To make up for the shortcomings of TSRMs in processing numerical values, the tools in TimeART are mainly chosen for handling the computation-intensive tasks. For example, scenario-aware tasks with the need of calculating statistical characteristics for long series or correlations among multiple series, and analytical tasks with the need of anomaly detection or forecasting. Considering the flexibility of TimeART, all the tools follow atomic designs, where only the basic computing needs are implemented without adding redundant functions. Merely retaining some necessary parameters, TSRM is allowed to autonomously decide how to utilize each atomic tool for implementing complex combinatorial analytical tasks through multiple iterations. Note that the forecasting and anomaly detection tools are two recently advanced lightweight time series foundation models, i.e., LightGTS~\citep{wang2025lightgts} and DADA~\citep{shentu2024towards}, which ensure both the efficiency and accuracy. The TimeART framework also supports adding self-defined tools for flexibility. All the details of the implemented tools can be found in Appendix~\ref{app: tool}.

\subsection{TimeToolBench}
\label{sec: timetoolbench}
With the help of TimeART, TSRMs can already utilize tools for reasoning. However, they still fall short in understanding the boundaries of tools' capabilities, which may cause suboptimal and conservative invocations. Intuitively, letting a senior data scientist demonstrate the standard trajectories, and leveraging TSRMs to learn the underlying principles is a feasible method, which has been widely applied in many other domains~\citep{gou2023tora,yao2022react}. To ensure the scale of the corpus, their approach often adopts powerful closed-source LLMs, 
to replace human experts.

\begin{figure}[t]
    \centering
\includegraphics[width=\linewidth]{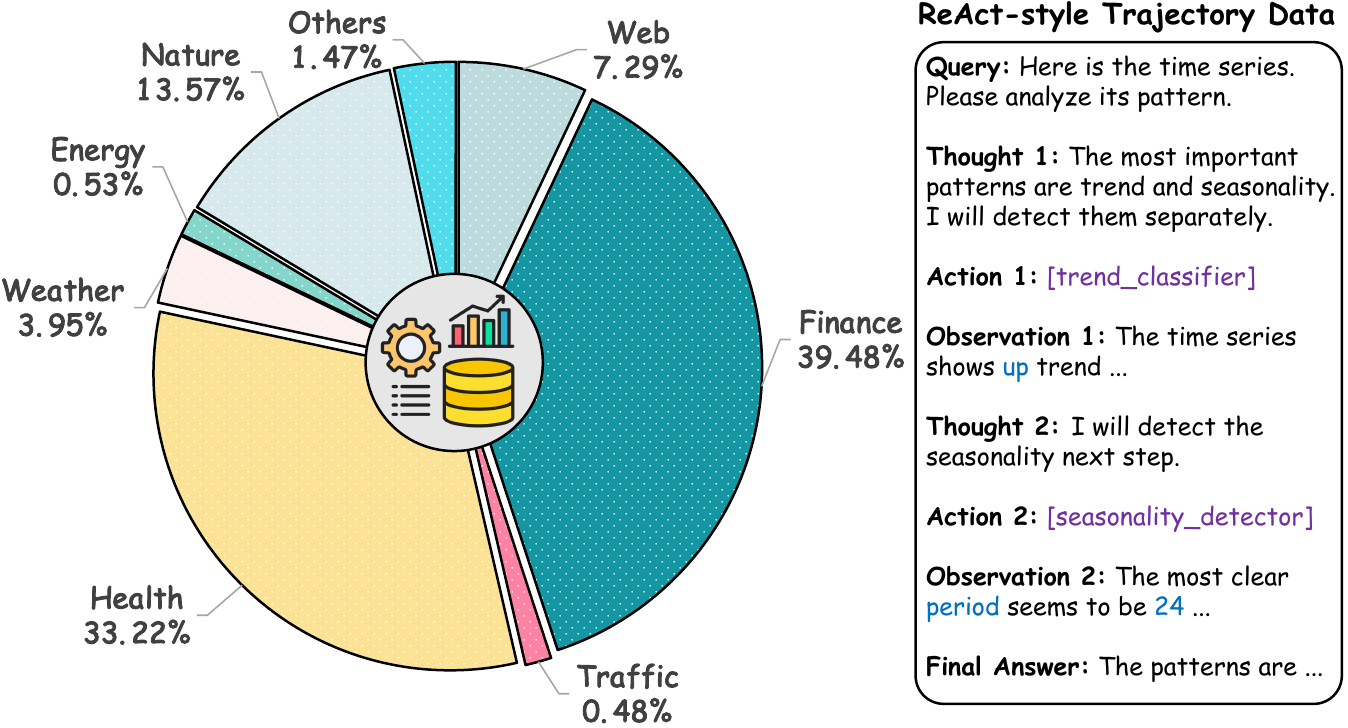}
    \caption{Ratios of data sources in TimeToolBench (100k), the ReAct-style training corpus for agentic time series reasoning.}
\label{fig: dataset}
\end{figure}

We collect and curate \textit{TimeToolBench}, which comprises over 100k tool-use trajectories with GPT-4o as the expert model, as shown in Figure~\ref{fig: dataset}. Since the tool-use trajectories are LLM artifacts, the original TSQA datasets used to generate them are from research teams~\citep{TimeMQA,cai2024timeseriesexam,chen2025mtbench,xie2024chatts}, which are from real-world applications or generated 

synthetically through human check, covering multiple domains shown in Figure~\ref{fig: dataset}. Compared with traditional chain-of-thought reasoning, this paradigm is able to generate both verbal reasoning traces and actions in an interleaved manner~\citep{yao2022react}. This enables LLMs to conduct dynamic reasoning to create, sustain, and adjust high-level action plans (reason to act). Simultaneously, it can interact with external environments to integrate supplementary information into the reasoning process (act to reason). Therefore, \textit{we also adopt the ReAct style in the inference phase of TimeART.}

During the collection of trajectories, we take some measures to ensure the quality of them--see Figure~\ref{fig: overview} (left). Given a TSQA sample $(\mathrm{Q}, y)$ with query $\mathrm{Q}$ and answer $y$, GPT-4o works like an expert and outputs a trajectory $\mathcal{T} := (\mathrm{Q}, (\mathrm{T}_1, \mathrm{A}_1, \mathrm{O}_1),\cdots,(\mathrm{T}_K, \mathrm{A}_K, \mathrm{O}_K), \mathrm{F})$. We first make coarse-grained answer check:
\begin{gather}
\text{Flag} = 
\begin{cases} 
1, & \text{with fixed options, } \mathrm{F} = y \\
1, & \text{open-ended, } \text{BERT-Score}(\mathrm{F}, y) > \sigma \\
0, & \text{otherwise}
\end{cases}
\end{gather}
For questions with fixed options, we ensure that the Final Answer F is completely consistent with y; for open-ended questions, since the answers are not fixed, we ensure that their semantic alignment with BERT-Score~\citep{reimers2019sentence} higher than a threshold $\sigma$. Then we filter out the trajectories with $\text{Flag}=0$. 
To preserve the quality of the reasoning process, we further verify the intermediate logical chains. Among tuples $(\mathrm{T}_k, \mathrm{A}_k, \mathrm{O}_k)$, the reasoning behaviors mainly occur at $\mathrm{T}_k \rightarrow \mathrm{A}_k$ and $\mathrm{O}_{k-1} \rightarrow \mathrm{T}_k$. Therefore, we randomly sample subsequences from the trajectory $\mathcal{T}$ like $(\mathrm{O}_{k-1}, \mathrm{T}_k, \mathrm{A}_k )$. To evaluate their qualities, we use three strong LLMs as judgers, i.e., DeepSeek-V3.2~\citep{liu2024deepseek}, Claude Opus 4.5, and Gemini 3 Pro; see Appendix~\ref{app: dataset}. For each LLM judger, we randomly sample a logical chain $(\mathrm{O}_{k-1}, \mathrm{T}_k, \mathrm{A}_k )$ from the trajectory $\mathcal{T}$, prompt it to evaluate whether the logical chain is reasonable. Supposed that there exists $M$ LLMs ($M=3$ here), the quality evaluation process can be formulated as:
\begin{gather}
\text{Flag} = \bigwedge_{m=1}^M \text{LLM}^m(\mathrm{O}_{k_m-1}, \mathrm{T}_{k_m}, \mathrm{A}_{k_m}),
\end{gather}
where $\text{LLM}^m(\mathrm{O}_{k_m-1}, \mathrm{T}_{k_m}, \mathrm{A}_{k_m})=1$ denotes that the $m$-th LLM gives positive evaluation to the logical chain $(\mathrm{O}_{k_m-1}, \mathrm{T}_{k_m}, \mathrm{A}_{k_m})$. Only if all LLMs give positive evaluations to the logical chains assigned to them, the $\text{Flag}=1$ and the trajectory is saved. This method can keep the diversity through using different LLMs and evaluated parts~\citep{badshah2025reference}, and keep strictness through requiring unanimous approval. 

\subsection{Training strategy}
\label{sec: training strategy}
To avoid the common dilemmas in behaviour cloning and reinforcement learning, we devise a training strategy to encourage TSRMs to learn from early experience and self-reflection. This strategy consists of four stages, covering the underlying principles of conventional behaviour cloning and reinforcement learning, and providing a more direct manner to teach TSRMs strategic tool-use. 

\begin{figure*}[t]
    \centering
\includegraphics[width=\linewidth]{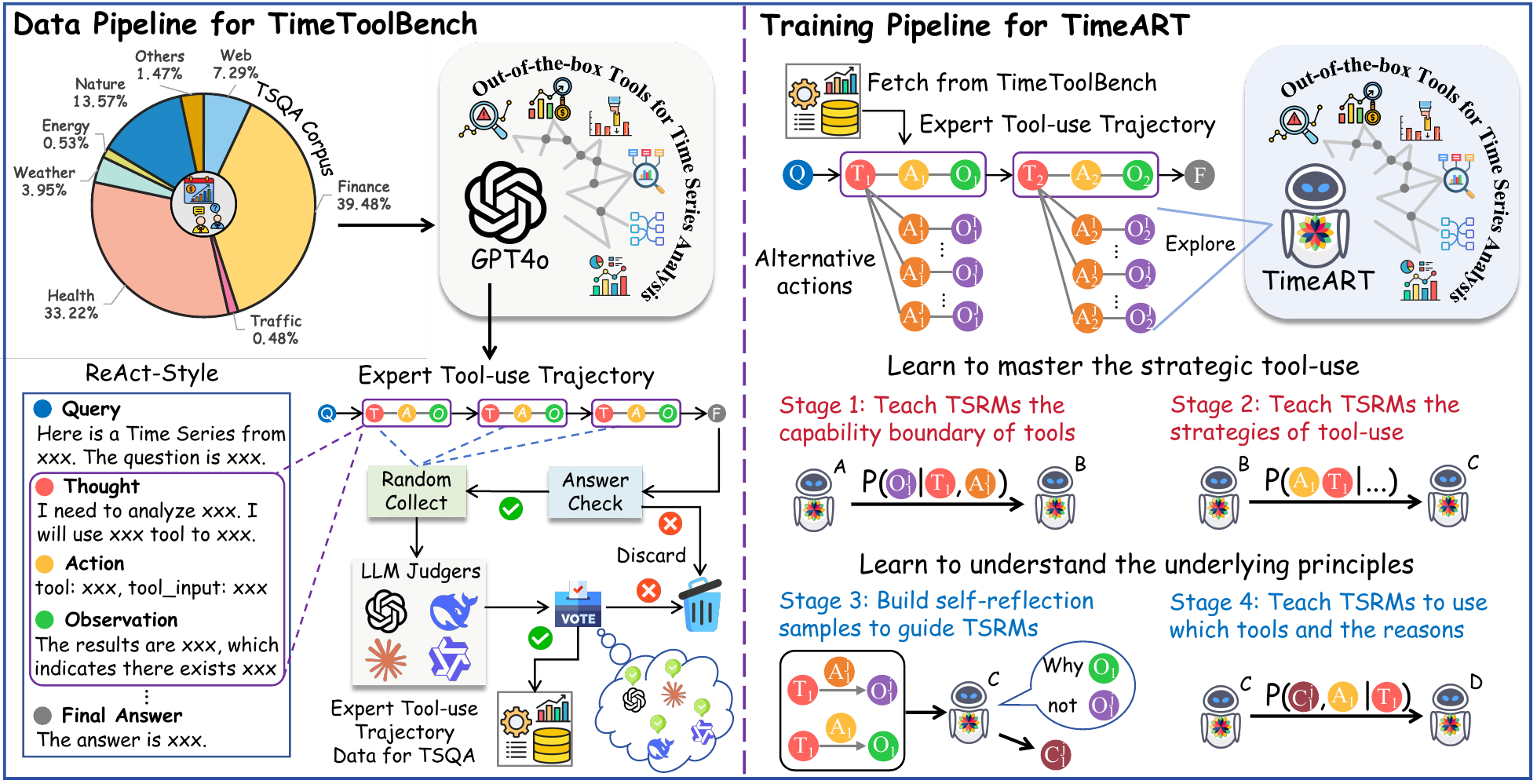}
    \caption{The overviews of the data pipeline for TimeToolBench (left), and the training pipeline for TimeART (right).}
\label{fig: overview}
\end{figure*}

\textbf{Stage 1: Teach TSRMs the capability boundary of tools.} Though the introductions of tools are organized as external constraints $\mathrm{E}$ to guide the invocations of TSRMs, they may not understand the actual capability of each tool. Under these circumstances, directly training TSRMs on TimeToolBench with behaviour cloning may lead to rigid imitation, thus hindering generalization. To tackle this problem, we first construct early experiences $\mathcal{D}_{exp}$ based on the collected trajectories. As shown in Figure~\ref{fig: overview} (right), considering a trajectory $\mathcal{T} := (\mathrm{Q}, (\mathrm{T}_1, \mathrm{A}_1, \mathrm{O}_1),\cdots,(\mathrm{T}_K, \mathrm{A}_K, \mathrm{O}_K), \mathrm{F})$ in TimeToolBench, we sample $J$ alternative tools for each Thought state $\mathrm{T}_k$ from the initial policy of the TSRM $p_\theta$: $\{\mathrm{A}_k^1, \cdots, \mathrm{A}_k^J\}$, then they lead to early experiences $\mathcal{D}_{exp}$: $\{(\mathrm{T}_k, \mathrm{A}_k^j, \mathrm{O}_k^j | \mathrm{S}_{1:3k-2})\}_{j=1,k=1}^{J,K}$, where $\mathrm{S}_{1:3k-2}$ are all previous states before $\mathrm{T}_k$. Then we train the TSRM $p_\theta$ to understand the capability boundary of tools:
\begin{gather}
    \mathcal{L}_1 = -\displaystyle\smashoperator{\sum_{(\mathrm{T}_k, \mathrm{A}_k^j, \mathrm{O}_k^j | \mathrm{S}_{1:3k-2})}}  \log p_\theta (\mathrm{O}_k^j | \mathrm{T}_k, \mathrm{A}_k^j, \mathrm{S}_{1:3k-2})
\end{gather}
This training objective allows TSRM $p_\theta$ to understand the transition rules with tool-use, and potential invalid and improper invocation experiences. Actually, it implicitly models a world model~\citep{gu2024your} by predicting the performance of tools in specific scenarios for agentic reasoning, which coarsely internalizes the dynamics into the TSRM.

\textbf{Stage 2: Teach TSRMs the strategies of tool-use.} After training the TSRM $p_\theta$ with objective $\mathcal{L}_1$, it can fully understand the functions of tools. In other words, Stage 1 serves as a good warmup phase, teaching the TSRM the basic use principles of TimeART. We then train TSRM for strategic tool-use:
\begin{gather}
    \mathcal{L}_2 = -\displaystyle\sum_{(\mathrm{T}_k,\mathrm{A}_k | \mathrm{S}_{1:3k-2})} \log p_\theta (\mathrm{A}_k, \mathrm{T}_k | \mathrm{S}_{1:3k-2})
\end{gather}
Note that though $\mathcal{L}_2$ is the conventional behaviour cloning objective, it guides the TSRM to learn reasoning and action rationales from the expert model after the warmup of Stage 1, which may mitigate the overfitting phenomena.

\textbf{Stage 3: Build self-reflection samples to guide TSRMs.} Then we build a self-reflection process to guide the TSRM to learn from its own exploratory outcomes. Specifically, we first use the trained TSRM $p_\theta$ to summarize the underlying explanations for tool invocation, which can provide richer, transferable supervision than expert actions alone. The process can be formulated as:
\begin{gather} 
    \text{For }(\mathrm{T}_k, \mathrm{A}_k^j, \mathrm{O}_k^j) \text{ and } (\mathrm{T}_k, \mathrm{A}_k, \mathrm{O}_k),\\
    \mathrm{C}_k^j = p_\theta(\text{Why }\mathrm{A}_k \text{ not } \mathrm{A}^j_k \text{ ?}),
\end{gather}
where $\mathrm{C}_k^j$ is the summarized explanation, explaining why the expert action $\mathrm{A}_k$ is preferable to the alternative action $\mathrm{A}_k^j$ based on the differences between their resulting observations $\mathrm{O}_k$ and $\mathrm{O}_k^j$. Thus the self-reflection dataset $\mathcal{D}_{ref} = \{(\mathrm{T}_k, \mathrm{A}_k^j, \mathrm{C}_k^j | \mathrm{S}_{1:3k-2})\}_{j=1,k=1}^{J,K}$ can be obtained. Note that the concrete prompt used to generate the self-reflection explanations are listed in Appendix~\ref{app: prompt}. 

\textbf{Stage 4: Teach TSRMs to use which tools and the reasons.} We then train the TSRM $p_\theta$ to jointly predict the explanation $\mathrm{C}^j_k$ and the expert action $\mathrm{A}_k$ conditioned on the current state:
\begin{gather}
    \mathcal{L}_3 = -\displaystyle\smashoperator{\sum_{(\mathrm{T}_k, \mathrm{A}_k^j, \mathrm{C}_k^j | \mathrm{S}_{1:3k-2})}} \log p_\theta(\mathrm{C}^j_k, \mathrm{A}_k, | \mathrm{T}_k, \mathrm{S}_{1:3k-2})
\end{gather}
Intuitively, this process teaches the TSRM to invoke proper tools based on the summarized criteria. Stages 3-4 can be regarded as self-distillation~\citep{DBLP:conf/nips/SchickDDRLHZCS23}, which extracts LLM’s own rationale, and encourages it to select tools based on it, thereby increasing the reliability. In practice, we mix the self-reflection dataset $\mathcal{D}_{ref}$ with the expert trajectories in TimeToolBench to construct the training samples for $\mathcal{L}_3$.

To sum up, learning from the Stage 1 -- 4 encourages TSRMs to move beyond simple imitations of expert trajectories. These stages construct an alternative way to mitigate the common dilemmas of conventional behaviour cloning and reinforcement learning in agentic time series reasoning, which teach TSRMs to master the strategic tool-use and understand the underlying principles, thus developing more generalizable decision criteria.

\section{Experiments}

\begin{table*}[t]
    \centering
     \resizebox{0.95\linewidth}{!}{
    \begin{tabular}{c|c|cc|cc|cc|cc|cc|cc}
    \toprule
        \multirow{3}{*}{Model} & \multirow{3}{*}{Type} & \multicolumn{4}{c|}{Stock Price Forecasting}  & \multicolumn{4}{c|}{Stock Indicator Forecasting} & \multicolumn{4}{c}{Temperature Forecasting}  \\ \cmidrule{3-14}
        ~ & ~ & \multicolumn{2}{c|}{7-day} & \multicolumn{2}{c|}{30-day} & \multicolumn{2}{c|}{7-day} & \multicolumn{2}{c|}{30-day} & \multicolumn{2}{c|}{7-day} & \multicolumn{2}{c}{14-day} \\ \cmidrule{3-14}
        ~ & ~ & MAE & MAPE & MAE & MAPE & MACD & BB & MACD & BB & MSE & MAE & MSE & MAE  \\ \midrule
        \rowcg
        \textit{GPT-4o} & closed & 1.596  & 2.544  & 2.338  & 3.520  & 0.365  & 1.082  & 0.897  & 2.068  & 17.550  & 3.110  & 40.430  & 4.490   \\ \midrule
        \textit{Gemini 3 Pro} & closed & 1.622  & 3.515  & 2.438  & 3.274  & 0.383  & 1.155  & 0.971  & 2.258  & 24.331  & 3.664  & 29.559  & 4.029   \\ \midrule
        \rowcg
        \textit{Claude Opus 4.5} & closed & 1.415  & 2.097  & 2.070  & 2.861  & 0.372  & 1.250  & 1.173  & 2.340  & 12.535  & 3.494  & 24.978  & 3.752   \\ \midrule
        \textit{DeepSeek-V3.2} & closed & 1.713  & 2.144  & 2.127  & 3.299  & 0.353  & 1.192  & 1.070  & 2.190  & 29.274  & 4.058  & 101.655  & 6.592  \\ \midrule
        \rowcg
        DeepSeek-R1 7B & open & 1.137 & 4.846 & \second{1.304} & 4.669 & 0.871  & 2.045  & 1.526  & 2.762  & 28.039  & 4.214  & 32.732 & 4.491  \\ \midrule
        ChatTS 7B & open & \second{0.866}  & \second{2.925}  & 1.339  & \second{3.071}  & 2.047  & 2.014  & 3.342  & \second{2.364}  & 19.512  & \second{3.407}  & \second{23.750} & \second{3.742}  \\ \midrule
        \rowcg
        ITFormer 7B & open & 0.983 & 2.928 & 1.403 & 3.133 & 0.368 & \second{1.226} & 1.482 & 2.632 & \second{16.000} & 3.983 & 27.384 & 4.025 \\\midrule
        Qwen3 8B & open & 1.106  & 5.441  & 1.313  & 5.241  & \second{0.359} & 2.038  & \second{0.887}  & 2.940  & 24.144  & 3.841  & 49.084 & 5.474   \\ \midrule
        \rowcr
        \textbf{TimeART} & open & \first{0.788}  & \first{1.634}  & \first{1.122}  & \first{3.054}  & \first{0.347}  & \first{1.032} & \first{0.862}  & \first{1.982}  & \first{4.021}  & \first{1.627}  & \first{5.026}  & \first{1.782}  \\ \bottomrule
    \end{tabular}}
    \caption{Results on Forecasting tasks in MTBench. Lower MSE, MAE, MAPE, MACD (MSE), and BB (MSE) values indicate better performance. \first{Red}: the best and \second{Blue}: the second best \textbf{among open-source models}; closed-source models are reported as a reference and are not ranked. We also present comparisons with baselines equipped with TimeART's framework in Table~\ref{tab:forecasting_tooluse}.}
    \label{tab: time series forecasting}
\end{table*}

\begin{table*}[!t]
    \centering
    \resizebox{0.95\linewidth}{!}{
    \begin{tabular}{c|c|cc|cc|cc|cc|c|c|c|c}
    \toprule
        \multirow{3}{*}{Model} & \multirow{3}{*}{Type} & \multicolumn{4}{c|}{Stock Trend Classification} & \multicolumn{4}{c|}{News-stock Correlation} & \multicolumn{2}{c|}{\makecell{News-finance \\ MCQA}} & \multicolumn{2}{c}{\makecell{Temperature Trend \\Classification}}   \\ \cmidrule{3-14}
        ~ & ~ & \multicolumn{2}{c|}{7-day} & \multicolumn{2}{c|}{30-day} & \multicolumn{2}{c|}{7-day} & \multicolumn{2}{c|}{30-day} & 7-day & 30-day & Past & Future  \\ \cmidrule{3-14}
        ~ & ~ & 3-way & 5-way & 3-way & 5-way & 3-way & 5-way & 3-way & 5-way & Acc & Acc & Acc & Acc  \\ \midrule
        \rowcg
        \textit{GPT-4o} & closed & 42.81  & 36.45  & 47.35  & 30.58  & 53.60  & 31.00  & 57.60  & 34.60  & 65.10  & 52.80   & 66.36  & 43.54   \\ \midrule
        \textit{Gemini 3 Pro} & closed & 47.11  & 41.64  & 44.78  & 29.71  & 51.85  & 26.39  & 59.43  & 34.82  & 63.38  & 50.29  & 56.91  & 51.54   \\ \midrule
        \rowcg
        \textit{Claude Opus 4.5} & closed & 44.85  & 33.48  & 52.00  & 31.60  & 50.39  & 29.01  & 57.99  & 34.21  & 75.59  & 61.15    & 59.77  & 57.12   \\ \midrule
        \textit{DeepSeek-V3.2} & closed & 45.05  & 35.44  & 48.41  & 29.59  & 49.98  & 27.01  & 57.53  & 35.06  & 77.47  & 69.50  & 26.56  & 25.15   \\ \midrule
        \rowcg
        DeepSeek-R1 7B & open & 27.53  & 17.11  & 46.46  & 26.06  & 57.84 & 32.99 & \second{53.49} & 31.01 & \first{90.40} & \first{89.12} & - & -  \\ \midrule
        ChatTS 7B & open & 19.71  & 16.09  & 19.38  & 13.46  & 46.44 & 24.44 & 37.60 & 22.67 & 85.91 & \second{84.47} & \second{60.80}  & 28.79   \\ \midrule
        \rowcg
        ITFormer 7B & open & \second{43.25} & \second{32.82} & \second{51.35} & 32.04 & 52.23 & 32.45 & 47.39 & \second{32.25} & 74.20 & 68.25 & 52.29 & \second{48.34}  \\\midrule
        Qwen3 8B & open & 37.55  & 32.71  & 44.20  & \second{32.50}  & \second{58.66}  & \second{35.64}  & 51.16  & 28.49  & 69.25  & 58.20  & 38.23  & 17.10   \\ \midrule
        \rowcr
        \textbf{TimeART} & open & \first{51.73}  & \first{45.07}  & \first{55.28}  & \first{41.25}  & \first{71.25}  & \first{42.36}  & \first{69.25}  & \first{40.52}  & \second{88.50}  & 78.50   & \first{76.22}  & \first{58.53}  \\ \bottomrule
    \end{tabular}}
    \caption{Results on Reasoning tasks in MTBench. Higher accuracy rates (Acc, 3-way, 5-way) indicate better performance. \first{Red}: the best and \second{Blue}: the second best \textbf{among open-source models}; closed-source models are reported as a reference and are not ranked. We also present comparisons with baselines equipped with TimeART's framework in Table~\ref{tab:reasoning_tooluse}.}
    \label{tab: time series reasoning}
\end{table*}

\begin{figure*}[t]
  \centering
  \begin{minipage}[b]{0.23\linewidth}
    \centering
    \subfloat[Understanding.]
    {\includegraphics[width=\linewidth]{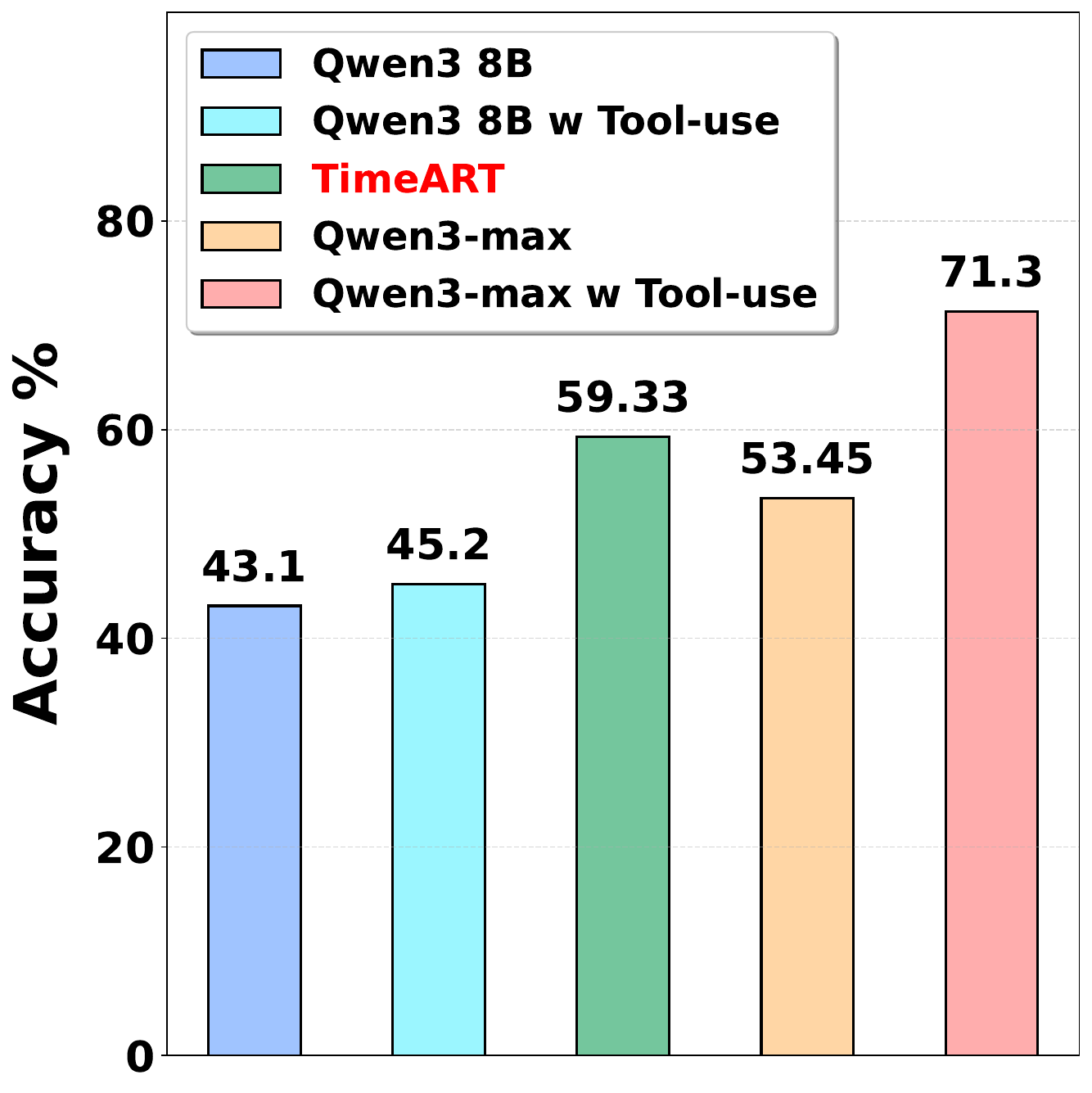}}
  \end{minipage}
  \hfill
  \begin{minipage}[b]{0.23\linewidth}
    \centering
    \subfloat[Perception.]
    {\includegraphics[width=\linewidth]{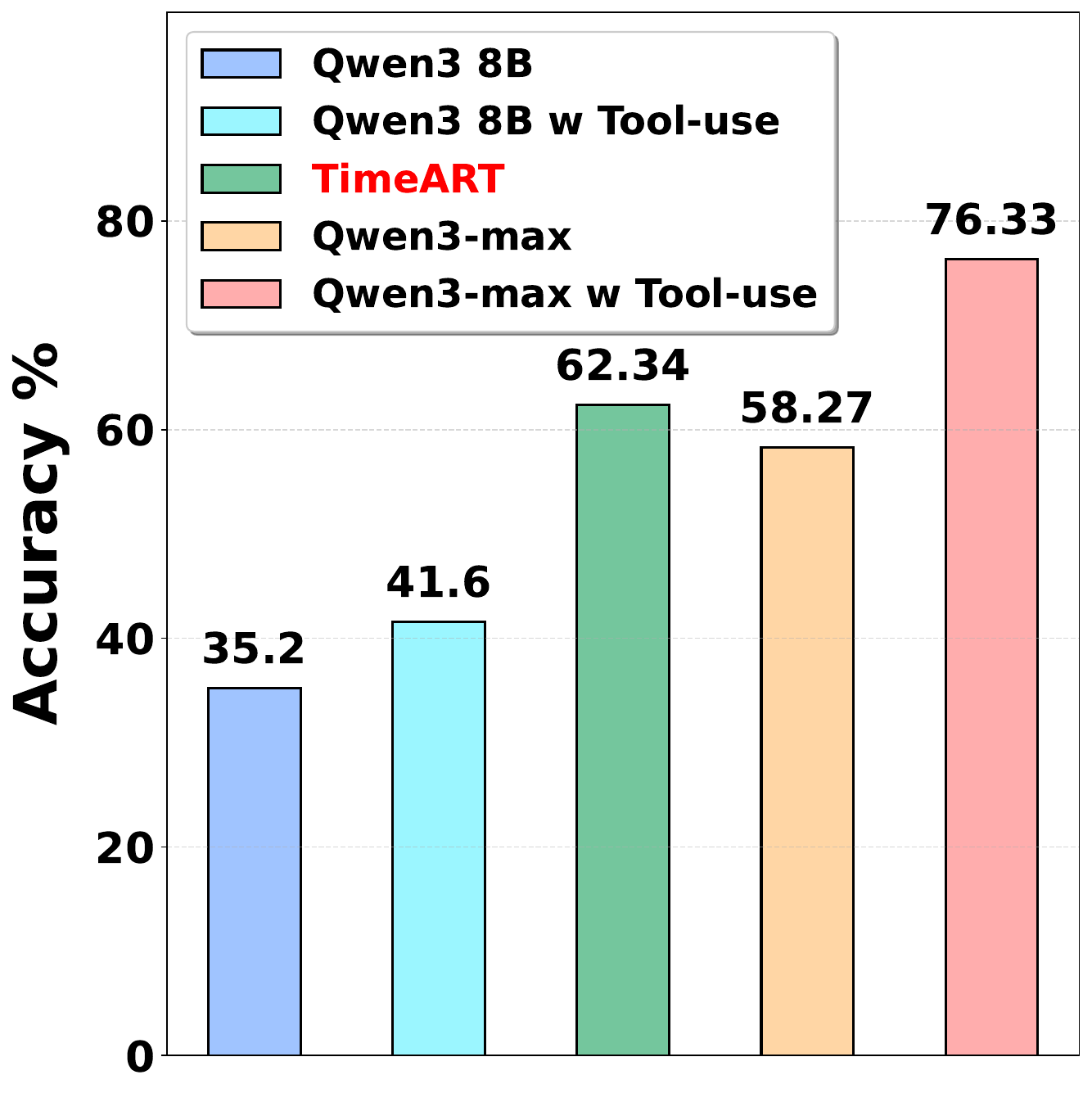}}
  \end{minipage}
  \hfill
  \begin{minipage}[b]{0.23\linewidth}
    \centering
    \subfloat[Reasoning.]
    {\includegraphics[width=\linewidth]{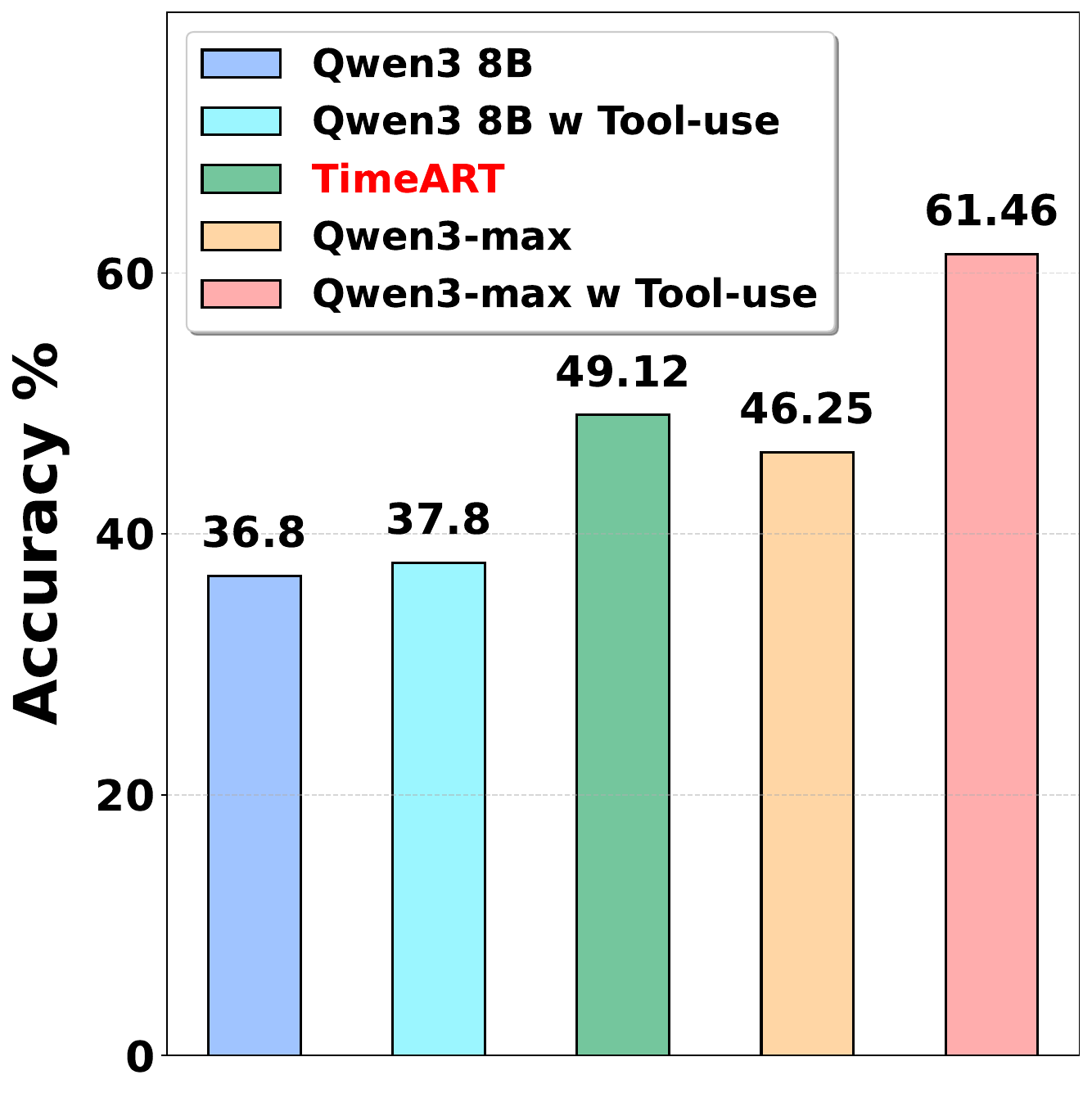}}
  \end{minipage}
  \hfill
  \begin{minipage}[b]{0.23\linewidth}
    \centering
    \subfloat[Estimation.]
    {\includegraphics[width=\linewidth]{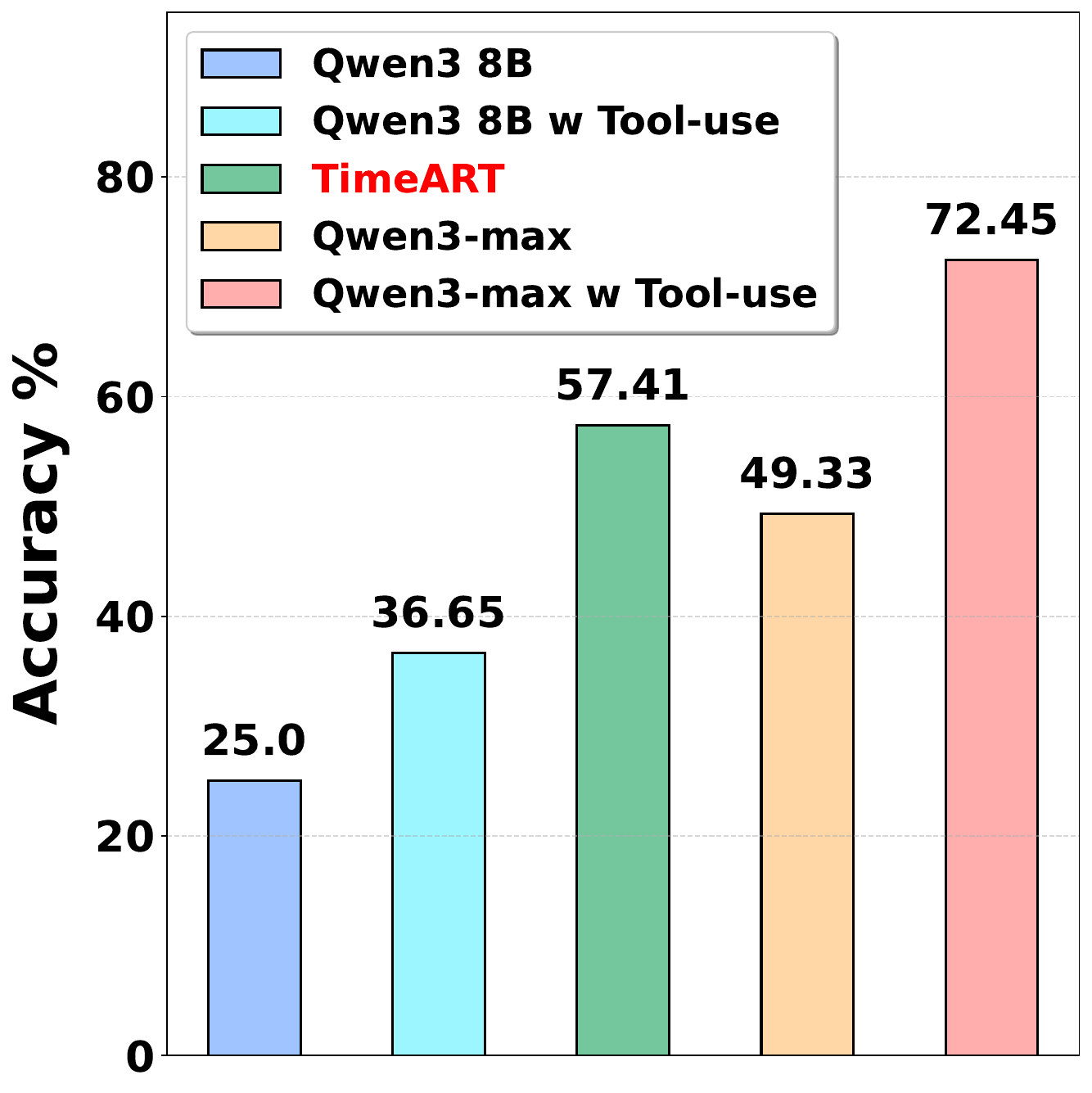}}
  \end{minipage}
  \caption{Ablations on the TimeART framework.}
  \label{fig: ablations on framework}
\end{figure*}

\subsection{Experimental Settings}
\textbf{Implementation.} We finetune an 8B Qwen3~\citep{yang2025qwen3} on TimeToolBench (Section~\ref{sec: timetoolbench}) with the proposed training strategy in Section~\ref{sec: training strategy}. During the training process, we adopt the LlamaFactory~\citep{zheng2024llamafactoryunifiedefficientfinetuning}, apply LoRA~\citep{hulora}, train the model with 8 NVIDIA 3090 GPUs. We then equip the trained Qwen3 with TimeART for evaluation. For simplification, \textit{we denote our method as TimeART in the following parts.}

\textbf{Benchmarks.} We evaluate TimeART on two well-recognized benchmarks MTBench~\citep{chen2025mtbench} and TimeMQA~\citep{TimeMQA}. Since part of the TimeToolBench is also constructed on source data in MTBench or TimeMQA, we remove them from the training set to avoid data leakage. 

\textbf{Evaluation Metrics.} We follow the evaluation protocols in TimeMQA and MTBench. For forecasting tasks like temperature forecasting and financial indicator forecasting, we follow the practice in MTBench to report the point-wise error, i.e., Mean Square Error (MSE), Mean Absolute Error (MAE), and Mean Absolute Percentage Error (MAPE). For TSQA tasks, we mainly adopt accuracy (Acc) as the metric. Note that there exist two kinds of accuracy (3-way and 5-way) in MTBench and we follow their settings.

\textbf{Baselines.} We compare TimeART against multiple closed-source and open-source models. For closed-source models, we choose GPT-4o~\citep{hurst2024gpt}, Gemini 3 Pro, DeepSeek-V3.2~\citep{liu2024deepseek}, Claude Opus 4.5, and Qwen3-max~\citep{yang2025qwen3}. For open-source models, we select DeepSeek-R1 7B, Llama3 8B, Mistral 7B, ChatTS 7B, ITFormer 7B, and Qwen3 8B. Note that not every baseline is applicable to every benchmark, so the model sets reported in Tables~\ref{tab: time series forecasting}--\ref{tab: TimeMQA} differ slightly.

\subsection{Main Results}
\textbf{Time Series Forecasting.} Table~\ref{tab: time series forecasting} presents forecasting results on stock and temperature data, where the tasks consider both textual and temporal information instead of the series alone. Following MTBench, Stock Price Forecasting has short-term (7-day input, 1-day output) and long-term (30-day input, 7-day output) settings reported with MAE and MAPE; Stock Indicator Forecasting targets Moving Average Convergence Divergence (MACD) and Upper Band of the Bollinger Bands (BB)~\citep{chen2025mtbench} reported with MSE; Temperature Forecasting has short-term (7-day input, 1-day output) and long-term (14-day input, 3-day output) settings reported with MSE and MAE. 

TimeART achieves state-of-the-art forecasting performance, with large improvements on Stock Price and Temperature Forecasting and a \textit{9\% -- 52\%} MAE reduction across all settings relative to the strongest open-source baseline of each setting. On Stock Indicator Forecasting, LLMs already perform well, which indicates these indicators are related to textual information; TimeART is still the best because it adaptively integrates tool results with the implicit information in texts.

\textbf{Time Series Reasoning.}
Table~\ref{tab: time series reasoning} presents reasoning results on MTBench, and Table~\ref{tab: TimeMQA} in Appendix~\ref{sec:more results} reports those on TimeMQA. The MTBench tasks (Stock Trend Classification, Temperature Trend Classification, News-stock Correlation, and News-finance MCQA) evaluate understanding and reasoning over domain texts and temporal correlations. For TimeMQA, we select the question-answer subsets and classify them into Understanding, Perception, Reasoning, and Estimation. Both are organized as multiple-choice questions, so we report accuracy; details are in Appendix~\ref{app: settings}.

On each single task, TimeART outperforms all open-source baselines a lot, demonstrating the effectiveness of strategic tool-use, and it stays competitive with closed-source GPT-4o and Qwen3-max using much fewer parameters. DeepSeek-R1 7B seems to encounter financial data during pre-training, performing decisively on financial reasoning tasks--see Table~\ref{tab: time series reasoning}--but it easily suffers over-thinking elsewhere, e.g, failure on Temperature Trend Classification and poor performance in Table~\ref{tab: TimeMQA}. Overall, TimeART shows sound tool-use strategies: it invokes the corresponding tools on analytical tasks, while on reasoning tasks it analyzes their outputs instead of showing blind faith in them. We make further model analyses in Section~\ref{sec: model analysis}.

\begin{figure*}[!t]
  \centering
  \begin{minipage}[b]{0.24\linewidth}
    \centering
    \subfloat[Understanding.]
    {\includegraphics[width=\linewidth]{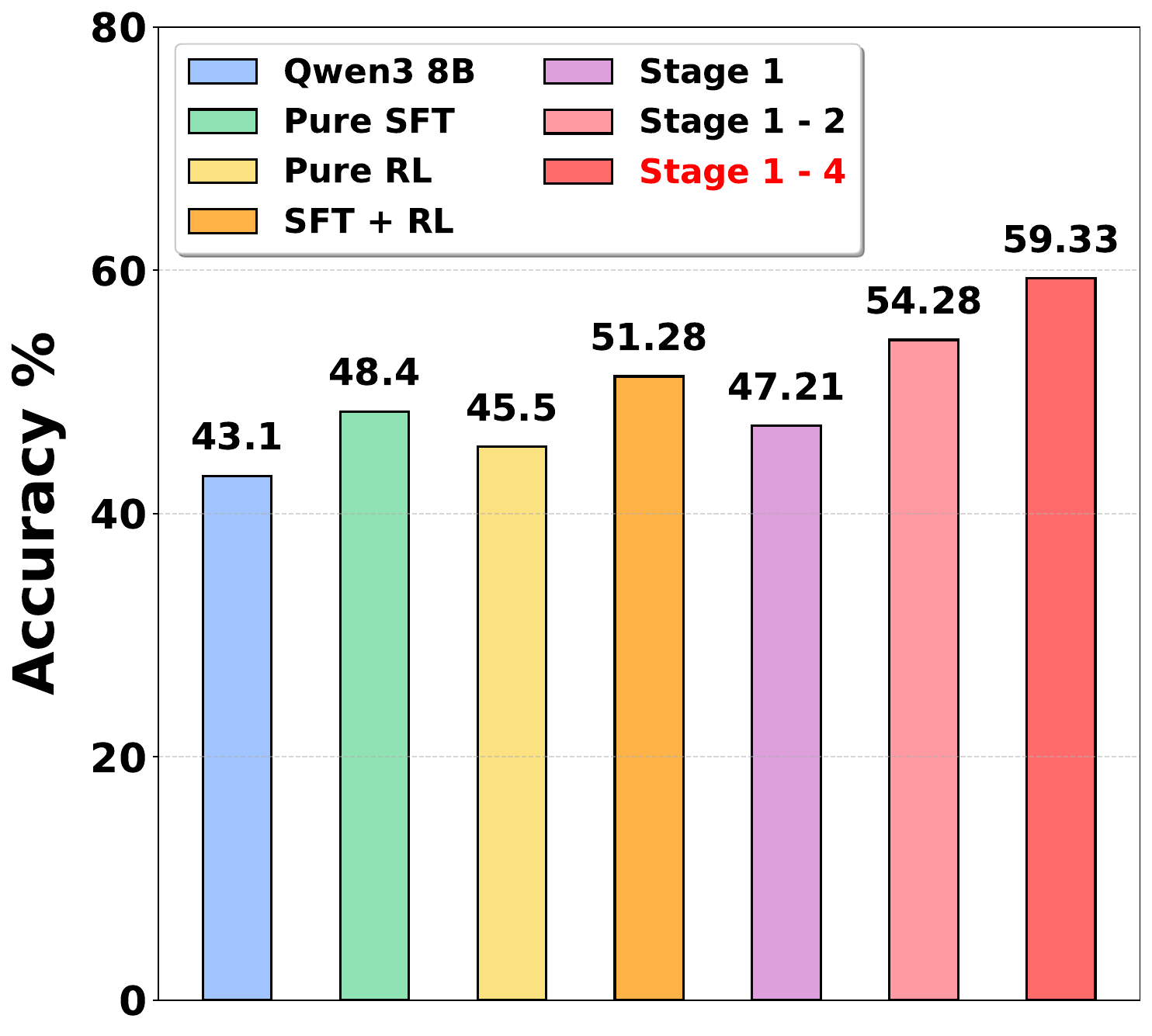}}
  \end{minipage}
  \hfill
  \begin{minipage}[b]{0.24\linewidth}
    \centering
    \subfloat[Perception.]
    {\includegraphics[width=\linewidth]{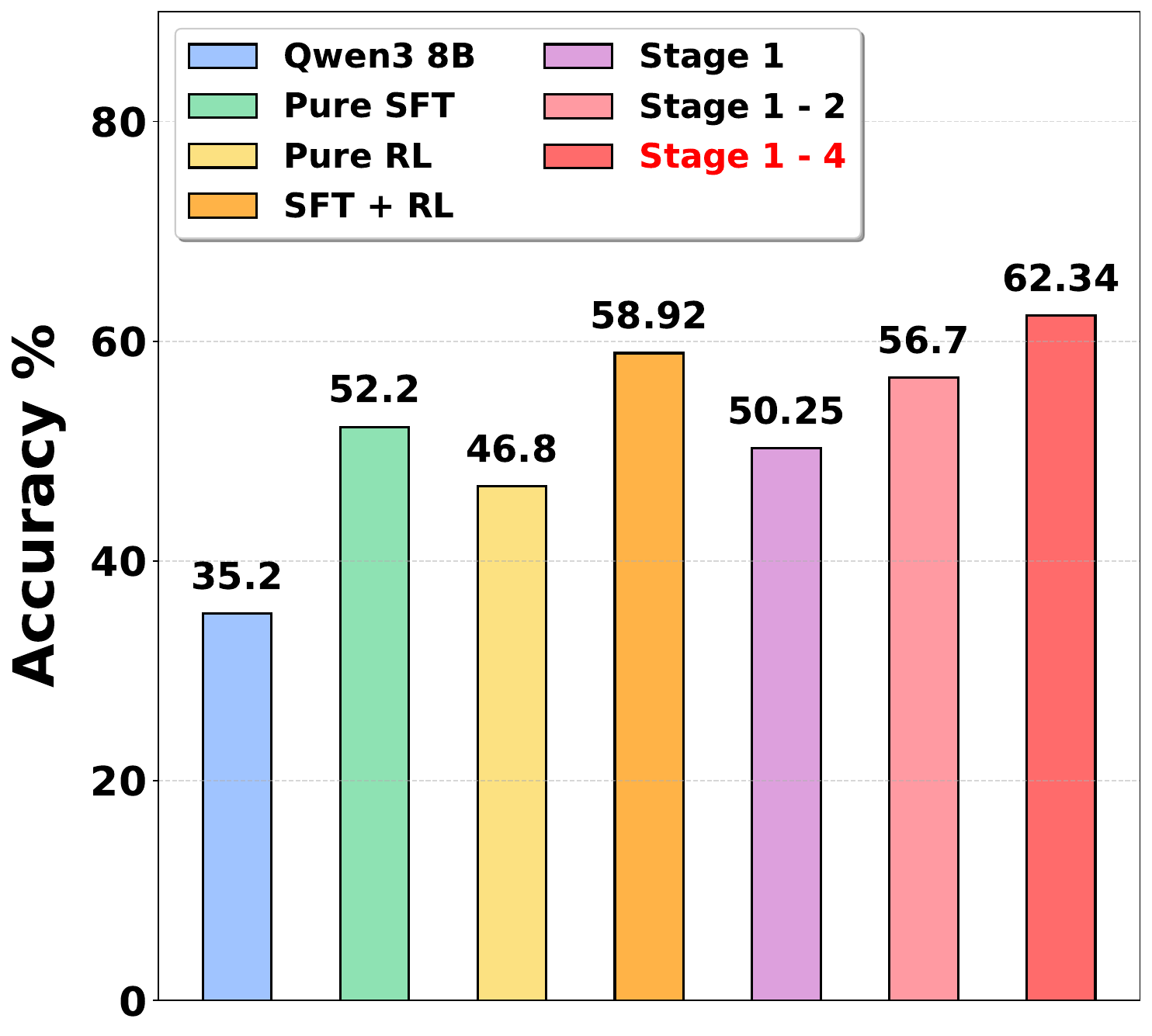}}
  \end{minipage}
  \hfill
  \begin{minipage}[b]{0.24\linewidth}
    \centering
    \subfloat[Reasoning.]
    {\includegraphics[width=\linewidth]{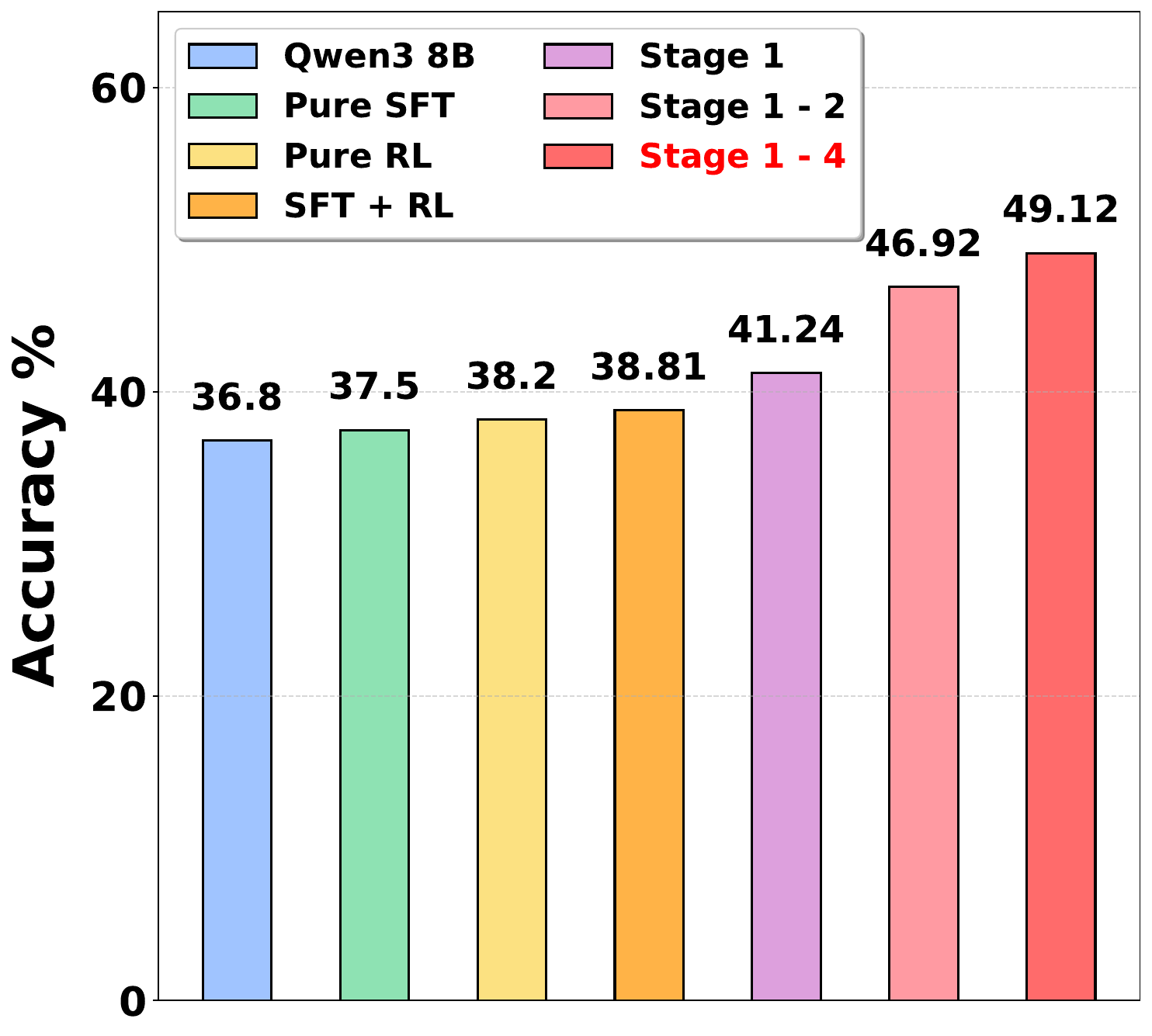}}
  \end{minipage}
  \hfill
  \begin{minipage}[b]{0.24\linewidth}
    \centering
    \subfloat[Estimation.]
    {\includegraphics[width=\linewidth]{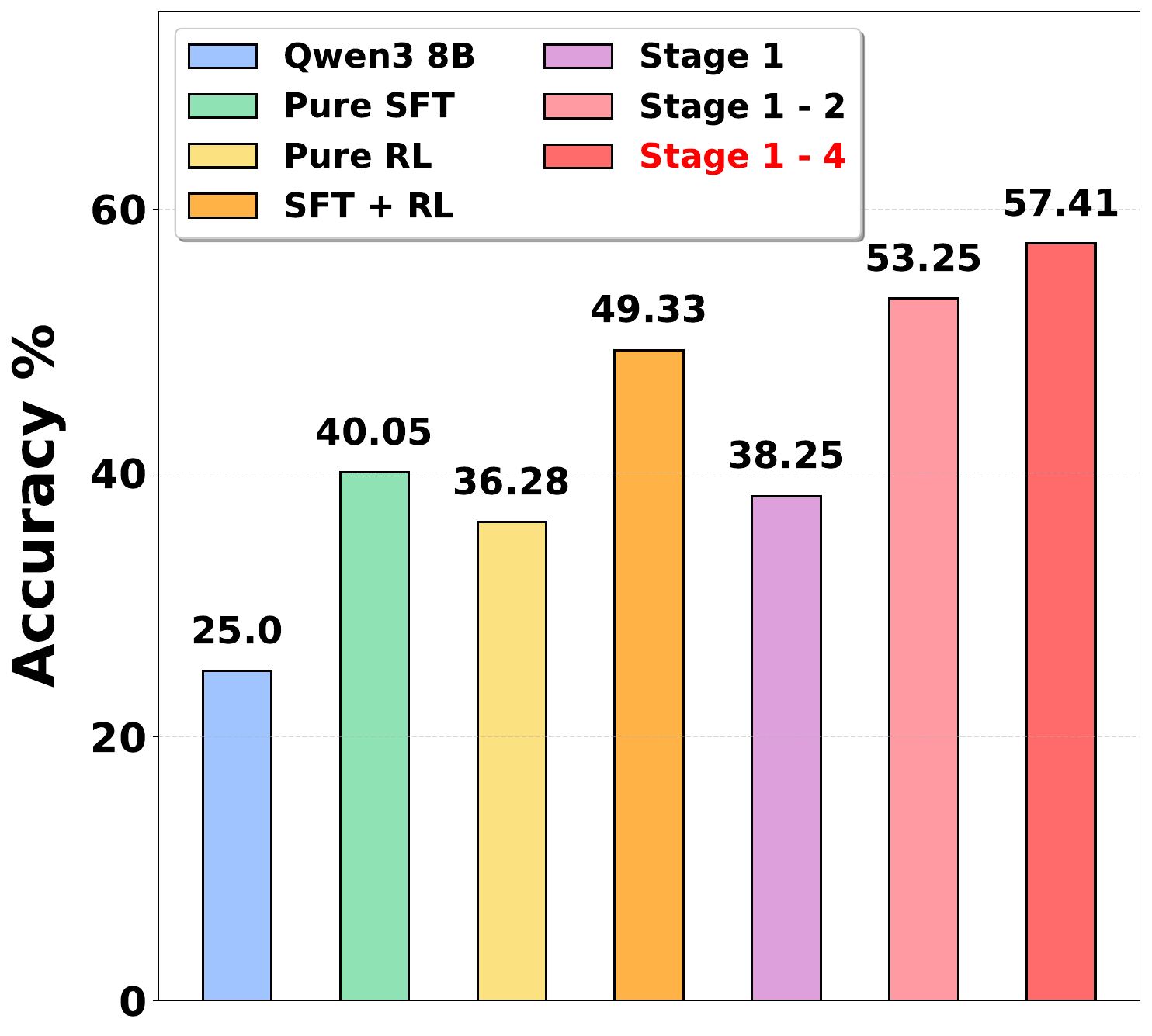}}
  \end{minipage}

  \caption{Ablations on training stages.}
  \label{fig: ablations on training}
\end{figure*}

\subsection{Model Analysis}
\label{sec: model analysis}

\textbf{Ablations on the framework.} Since TimeART can be treated as a general tool-use framework, we evaluate five variants on the four TSQA categories from TimeMQA (Figure~\ref{fig: ablations on framework}): original Qwen3 8B, Qwen3 8B with Tool-use (the original model in TimeART), our trained Qwen3 8B with TimeART, Qwen3-max, and Qwen3-max with Tool-use. We have following key observations: 1) Tool use is important to time series reasoning, since both Qwen3 8B and Qwen3-max improve directly once equipped with the Tool-use in TimeART; 2) a cascading effect appears where stronger LLMs gain more, as TimeART improves Qwen3-max more than Qwen3 8B on all four tasks; 3) TimeToolBench is effective for finetuning small-scale LLMs, as the finetuned Qwen3 8B acquires the strategic tool-use ability and, with TimeART, outperforms the closed-source Qwen3-max without tool-use.

\textbf{Ablations on training stages.} We also validate the training strategy, in which Stage 1 and Stage 2 are independent while Stages 3 and 4 are coherent. On the same four categories, Figure~\ref{fig: ablations on training} compares the original Qwen3 8B, conventional (SFT + RL (GRPO)), only Stage 1, only Stages 1 and 2, and all the training stages (ours). We have the following key conclusions: 1) Stage 1 is important, as it consistently improves all TSQA tasks and even outperforms conventional (SFT + RL) on the reasoning task; tool capability boundaries are thus key information that Stage 1 helps TSRMs internalize like a world model; 2) Stage 1 mitigates the low generalization dilemma in SFT (Stage 2), since Stages 1 and 2 alone outperform conventional (SFT + RL) in most cases: the taught capability boundaries preserve prior knowledge and make Stage 2 beyond simple imitations, keeping the general tool-use principles from converging to conservative decisions; 3) Learning from self-reflection (Stage 3 + 4) is also vital, as training on all stages is consistently the best. These two stages replace fixed manually-designed preferences (reward models) with flexible self-reflected internal logic, thus boosting self-evolution of agents.

\section{Related Works}
\label{sec: related works}
Previous studies mainly utilize Large Language Models in conventional analytical tasks such as forecasting and anomaly detection, aiming to endow time series models with generalization capabilities transferred from LLMs and thus alleviate the domain-specific limitations that arise when large-scale time series corpus is scarce. However, some studies~\citep{zhou2024can,tan2024language} point out that LLMs do not achieve significant performance compared to deep time series models but possess larger overheads.

LLMs are also suitable for tasks with high requirements for cross-modality and interpretability. Therefore, recent studies like ChatTS~\citep{xie2024chatts}, TimeOmni-1~\citep{guan2025timeomni1incentivizingcomplexreasoning}, ITformer~\citep{leiitformer}, and TRACE~\citep{anonymous2026trace}, enhance the inherent reasoning capability of LLMs in TSQA tasks, helping data scientists solve problems from vertical domains such as finance, weather, and aircraft engine~\citep{leiitformer}. However, most efforts are limited to improving the ability of LLMs themselves, neglecting that numerical ability is a weakness of LLMs. In this work, we introduce TimeART to systematically explore agentic time series reasoning, combining the advantages of analytical tools in numerical processing with the strong reasoning capability of LLMs.

Recent studies on time series agents mainly focus on conventional tasks that are already well solved by conventional methods, leveraging the decision-making capability of LLM-based agents or multi-agents to pursue automation or few-shot inference. Though some studies focus on time series reasoning agents~\citep{liu2025tsagenttimeseriesreasoning,ye2025beyond}, they put emphasis on designing a fixed procedure or do not fully exploit the diversity of external analytical tools, and do not well adapt LLMs to the time series domain through training. In this work, we incorporate TimeART with strong tools, and propose TimeToolBench to finetune TSRMs with a novel training strategy. Representative methods, benchmarks, and further discussion are provided in Appendix~\ref{sec:further related}.

\section{Conclusion}
We introduce TimeART, an agentic framework that equips TSRMs with out-of-the-box tools, plus TimeToolBench, a \textit{100k} trajectory corpus and a four-stage strategy trained on expert and self-generated data. An 8B TSRM with TimeART reaches state of the art on multiple benchmarks.

\section*{Limitations}
TimeART performs strongly across a range of Time Series Question Answering tasks, and our experiments point to several directions along which it can be taken further. First, TimeART builds on a curated set of external analytical tools, so reaching domains or question types beyond their current coverage benefits from extending the toolset, which involves a lightweight additional Stage 1 pass so that the capability boundary of a new tool is learned, as discussed in Appendix~\ref{app: tool}. Second, our evaluation centers on an 8B-scale TSRM, and characterizing how the observed gains scale to substantially smaller or larger backbones is a natural next step. Third, extremely long or high-dimensional multivariate series place additional demands on efficiency and context retention. Appendix~\ref{app:limitations} discusses these points in more detail, together with the attribution of the gains and the scope of the evaluation benchmarks. We leave these directions to future work.

\section*{Ethics Statement}
This work only employs publicly available datasets and commonly used open-source tools and frameworks, and we foresee no direct negative societal impacts beyond those generally associated with LLM-based systems. All external assets, including open-source code, public datasets, and pre-trained models, are properly cited and used in accordance with their licenses. We use LLMs only for grammar checking and polishing of our manually written content.

\bibliography{reference}

\clearpage
\appendix
\tcbset{compress page=none}

\section{More Experimental Results}
\label{sec:more results}
\subsection{Full results on TimeMQA}
Table~\ref{tab: TimeMQA} reports the accuracy of TimeART and all baselines on the four TimeMQA categories, which complements the MTBench results in Table~\ref{tab: time series reasoning}.

\begin{table}[t]
    \centering
    \resizebox{\linewidth}{!}{
    \begin{tabular}{c|c|c|c|c|c}
    \toprule
        \rowcb
        Model & Type & Understanding & Perception & Reasoning & Estimation \\ \midrule
        \rowcg
        \textit{GPT-4o} & closed & 50.86  & 69.65  & 50.00  & 66.66  \\ \midrule     
        \textit{Qwen3-max} & closed &53.45 &58.27 &46.25 &49.33  \\ \midrule   
        \rowcg
        DeepSeek-R1 7B & open  & 7.76  & 12.69  & 16.89  & 7.41  \\ \midrule
        
        Llama3 8B & open  & 33.62  & 36.27  & 27.03  & 25.92  \\ \midrule
        \rowcg
        ChatTS 7B & open  & 25.00  & 28.68  & \second{38.51}  & 12.03  \\ \midrule
        ITFormer 7B & open & 40.52 & \second{45.24} & 30.40 & \second{44.44}\\\midrule
        \rowcg
        Qwen3 8B & open  & \second{43.10}  & 35.20  & 36.80  & 25.00  \\ \midrule
        \makecell{TimeMQA + \\ Llama3 8B} & open&28.45 &36.27 &22.97 &40.74 \\ \midrule
        \rowcg
        \makecell{TimeMQA + \\ Qwen2.5 7B} &open &10.34 &24.41 &13.51 &26.85 \\ \midrule
        \makecell{TimeMQA + \\ Mistral 7B} & open  & 18.96  & 26.20  & 12.83  & 
        28.70  \\ \midrule
        \rowcr
        \textbf{TimeART} & open & \first{59.33}  & \first{62.34}  & \first{49.12}  & \first{57.41} \\ \bottomrule
    \end{tabular}}
    \caption{Accuracy rates on tasks in TimeMQA. Higher values indicate better performance. \first{Red}: the best and \second{Blue}: the second best \textbf{among open-source models}; closed-source models are reported as a reference and are not ranked. }
    \label{tab: TimeMQA}
\end{table}

\subsection{More comparisons with baselines equipped with TimeART's tool-use framework}

To further demonstrate the effectiveness of our proposed training strategy, we equip all baselines with the tool-use framework of TimeART and allow few-shot contextual prompts to eliminate the performance gains brought by this framework. In terms of baselines, we further consider an even more powerful closed-source model, GPT-5.2. We also introduce DeepAnalyze~\citep{DBLP:journals/corr/abs-2510-16872}, an open-source data analytical agent as another baseline. The experimental results on MTBench show that TimeART still achieves the optimal performance in most cases, significantly outperforming untrained open-source models, which verifies the effectiveness of our training strategy. As shown in Table~\ref{tab:forecasting_tooluse}--\ref{tab:reasoning_tooluse}, it is observed that when compared with closed-source models enhanced by few-shot prompts, TimeART also exhibits competitive advantages and achieves comparable or even better performance.

\begin{table*}[t]
    \centering
     \resizebox{1\linewidth}{!}{
    \begin{tabular}{c|c|cc|cc|cc|cc|cc|cc}
    \toprule
        \multirow{3}{*}{Model} & \multirow{3}{*}{Type} & \multicolumn{4}{c|}{Stock Price Forecasting}  & \multicolumn{4}{c|}{Stock Indicator Forecasting} & \multicolumn{4}{c}{Temperature Forecasting}  \\ \cmidrule{3-14}
        ~ & ~ & \multicolumn{2}{c|}{7-day} & \multicolumn{2}{c|}{30-day} & \multicolumn{2}{c|}{7-day} & \multicolumn{2}{c|}{30-day} & \multicolumn{2}{c|}{7-day} & \multicolumn{2}{c}{14-day} \\ \cmidrule{3-14}
        ~ & ~ & MAE & MAPE & MAE & MAPE & MACD & BB & MACD & BB & MSE & MAE & MSE & MAE  \\ \midrule
        \rowcg
        \textit{GPT-4o} & closed 
        & 0.805 & 1.720 & 1.142 & 3.080 
        & 0.352 & 1.052 & 0.870 & 2.006 
        & 4.145 & 1.665 & 5.170 & 1.835 \\ \midrule
        
        \textit{GPT-5.2} & closed 
        & 0.802 & 1.650 & 1.108 & 3.070 
        & 0.350 & 1.048 & 0.868 & 2.004 
        & 4.125 & 1.658 & 5.145 & 1.822 \\ \midrule
        
        \rowcg
        \textit{Gemini 3 Pro} & closed 
        & 0.795 & 1.616 & 1.133 & 3.066 
        & 0.3508 & 1.046 & 0.8654 & 1.994 
        & 4.074 & 1.652 & 5.052 & 1.757 \\ \midrule
        
        \textit{Qwen3-max} & closed 
        & 0.810 & 1.780 & 1.150 & 3.090 
        & 0.345 & 1.060 & 0.872 & 2.018 
        & 4.110 & 1.665 & 5.150 & 1.850 \\
        \midrule
        
        \rowcg
        DeepAnalyze 8B & open 
        & 0.862 & \second{2.180} & 1.218 & 3.355 
        & 0.392 & \second{1.278} & 1.018 & \second{2.168} 
        & \second{7.850} & \second{2.060} & \second{10.920} & \second{2.455} \\ \midrule
        
        DeepSeek-R1 7B & open 
        & 0.956 & 2.905 & 1.246 & 3.820 
        & 0.438 & 1.384 & 1.086 & 2.286 
        & 9.960 & 2.435 & 14.860 & 2.925 \\ \midrule
        
        \rowcg
        ChatTS 7B & open 
        & \second{0.838} & 2.465 & 1.232 & \second{3.285} 
        & 1.612 & 1.492 & 2.215 & 2.185 
        & 8.320 & 2.185 & 12.240 & 2.585 \\ \midrule
        
        Qwen3 8B & open 
        & 0.905 & 3.735 & \second{1.188} & 3.875 
        & \second{0.356} & 1.418 & \second{0.876} & 2.386 
        & 8.980 & 2.320 & 16.720 & 3.045 \\ \midrule
        
        \rowcr
        \textbf{TimeART 8B} & open 
        & \first{0.788}  & \first{1.634}  
        & \first{1.122}  & \first{3.054}  
        & \first{0.347}  & \first{1.032} 
        & \first{0.862}  & \first{1.982}  
        & \first{4.021}  & \first{1.627}  
        & \first{5.026}  & \first{1.782}  \\ 
        \bottomrule
    \end{tabular}}
    \caption{Results on Forecasting tasks in MTBench. \textbf{Note that we equip all the baselines with TimeART's tool-use framework, and they are prompted with few-shot examples.} Lower MSE, MAE, MAPE, MACD (MSE), and BB (MSE) values indicate better performance. \first{Red}: the best and \second{Blue}: the second best \textbf{among open-source models}; closed-source models are reported as a reference and are not ranked.}
    \label{tab:forecasting_tooluse}
\end{table*}

\begin{table*}[t]
    \centering
    \resizebox{1\linewidth}{!}{
    \begin{tabular}{c|c|cc|cc|cc|cc|c|c|c|c}
    \toprule
        \multirow{3}{*}{Model} & \multirow{3}{*}{Type} 
        & \multicolumn{4}{c|}{Stock Trend Classification} 
        & \multicolumn{4}{c|}{News-stock Correlation} 
        & \multicolumn{2}{c|}{\makecell{News-finance \\ MCQA}} 
        & \multicolumn{2}{c}{\makecell{Temperature Trend \\Classification}}   \\ 
        \cmidrule{3-14}
        ~ & ~ 
        & \multicolumn{2}{c|}{7-day} 
        & \multicolumn{2}{c|}{30-day} 
        & \multicolumn{2}{c|}{7-day} 
        & \multicolumn{2}{c|}{30-day} 
        & 7-day & 30-day & Past & Future  \\ 
        \cmidrule{3-14}
        ~ & ~ 
        & 3-way & 5-way 
        & 3-way & 5-way 
        & 3-way & 5-way 
        & 3-way & 5-way 
        & Acc & Acc & Acc & Acc  \\ 
        \midrule
        
        \rowcg
        \textit{GPT-4o} & closed 
        & 49.85 & 42.76 
        & 53.40 & 38.15 
        & 68.45 & 40.62 
        & 66.84 & 38.92 
        & 86.75 & 75.40 
        & 73.85 & 55.92 \\ 
        \midrule
        
        \textit{GPT-5.2} & closed 
        & 52.38 & 45.62 
        & 54.86 & 40.72 
        & 70.82 & 41.95 
        & 68.74 & 40.18 
        & 90.65 & 81.25 
        & 75.90 & 58.10 \\ 
        \midrule
        
        \rowcg
        \textit{Gemini 3 Pro} & closed 
        & 51.80 & 44.85 
        & 54.45 & 39.86 
        & 70.35 & 41.56 
        & 68.28 & 39.84 
        & 89.72 & 79.85 
        & 75.35 & 57.62 \\ 
        \midrule
        
        \textit{Qwen3-max} & closed 
        & 50.76 & 43.96 
        & 53.92 & 39.20 
        & 69.88 & 41.20 
        & 67.95 & 39.50 
        & 89.10 & 79.20 
        & 75.10 & 57.30 \\
        \midrule
        
        \rowcg
        DeepAnalyze 8B & open 
        & \second{47.82} & \second{39.86} 
        & \second{52.12} & \second{36.74} 
        & \second{66.10} & \second{38.24} 
        & \second{64.52} & \second{36.85} 
        & 84.80 & 74.52 
        & \second{71.18} & \second{53.62} \\ 
        \midrule
        
        DeepSeek-R1 7B & open 
        & 39.45 & 26.92 
        & 49.35 & 32.84 
        & 62.55 & 35.78 
        & 58.24 & 33.82 
        & \second{88.20} & \second{78.20} 
        & 63.05 & 48.90 \\ 
        \midrule
        
        \rowcg
        ChatTS 7B & open 
        & 34.20 & 25.85 
        & 38.06 & 27.12 
        & 55.02 & 30.35 
        & 50.15 & 28.86 
        & 87.05 & 77.92 
        & 69.10 & 43.55 \\ 
        \midrule
        
        Qwen3 8B & open 
        & 45.62 & 39.05 
        & 50.24 & 35.08 
        & 64.85 & 37.02 
        & 61.48 & 34.76 
        & 81.50 & 70.62 
        & 59.80 & 39.20 \\ 
        \midrule
        
        \rowcr
        \textbf{TimeART 8B} & open 
        & \first{51.73} & \first{45.07} 
        & \first{55.28} & \first{41.25} 
        & \first{71.25} & \first{42.36} 
        & \first{69.25} & \first{40.52} 
        & \first{88.50} & \first{78.50} 
        & \first{76.22} & \first{58.53} \\ 
        \bottomrule
    \end{tabular}}
    \caption{Results on Reasoning tasks in MTBench. \textbf{Note that we equip all the baselines with TimeART's tool-use framework, and they are prompted with few-shot examples.} Higher accuracy rates (Acc, 3-way, 5-way) indicate better performance. \first{Red}: the best and \second{Blue}: the second best \textbf{among open-source models}; closed-source models are reported as a reference and are not ranked.}
    \label{tab:reasoning_tooluse}
\end{table*}

\subsection{Attribution of improvements: analytical tools versus the training strategy}
\label{sec:attribution}

A natural question is how much of TimeART's improvement stems from the analytical tools themselves rather than from the proposed four-stage training. To isolate the two factors, we compare five variants that share the identical 21-tool framework, the same tool implementations, and the same evaluation protocol, and differ only in the policy deciding which tool to invoke: (a) \textit{random selection}, which samples a tool uniformly at each step; (b) a \textit{fixed heuristic policy}, which applies a deterministic mapping from question type to tool (forecasting questions to \texttt{forecaster}, trend questions to \texttt{trend\_classifier}, anomaly questions to \texttt{anomaly\_detection}, and so on); (c) \textit{Qwen3 8B with tools, zero-shot}; (d) \textit{Qwen3 8B with tools, few-shot}, the strongest policy obtainable without training; and (e) TimeART, i.e., Qwen3 8B trained with all four stages.

Throughout Appendix~\ref{sec:more results}, aggregate scores are defined so that they remain directly comparable with the tables and figures of the main text. TimeMQA accuracy is the macro-average over the four categories (Understanding, Perception, Reasoning, and Estimation) of Table~\ref{tab: TimeMQA}; MTBench reasoning accuracy is the macro-average over the twelve task settings of Table~\ref{tab: time series reasoning}; and forecasting MAE is the mean of the four MAE columns of Table~\ref{tab: time series forecasting}, i.e., Stock Price Forecasting at 7 and 30 days and Temperature Forecasting at 7 and 14 days. Every row of Table~\ref{tab:attribution} is thereby traceable: the zero-shot variant aggregates the \textit{Qwen3 8B with Tool-use} bars of Figure~\ref{fig: ablations on framework}, the few-shot variant aggregates the Qwen3 8B rows of Tables~\ref{tab:forecasting_tooluse} and~\ref{tab:reasoning_tooluse}, and the TimeART row aggregates Tables~\ref{tab: time series forecasting}, \ref{tab: time series reasoning}, and~\ref{tab: TimeMQA}.

Table~\ref{tab:attribution} reports the results. Tool access alone already helps, since the fixed heuristic policy clearly outperforms random selection and both untrained Qwen3 8B variants improve once the tools become available. However, the learned policy adds a further 15.85 points on TimeMQA and 8.27 points on MTBench reasoning over the strongest untrained variant, and reduces forecasting MAE from 1.86 to 1.33. This margin cannot be explained by tool access, because every variant uses the same tools, and therefore attributes the dominant part of the improvement to the training strategy. Since training also exposes the model to 75.4k domain trajectories, the per-stage ablation of Figure~\ref{fig: ablations on training}, which holds both the tools and the data fixed, complements this experiment by separating the contributions of the individual stages.

\begin{table}[t]
    \centering
    \resizebox{\linewidth}{!}{
    \begin{tabular}{l|c|c|c}
    \toprule
        \rowcb
        Tool-use policy & \makecell{TimeMQA \\ Acc} & \makecell{MTBench \\ Reason. Acc} & \makecell{MTBench \\ Forec. MAE} \\ \midrule
        \rowcg
        Random selection & 28.40 & 35.24 & 3.09 \\ \midrule
        Fixed heuristic policy & 38.70 & 46.54 & 1.93 \\ \midrule
        \rowcg
        Qwen3 8B + tools (zero-shot) & 40.31 & 43.84 & 2.15 \\ \midrule
        Qwen3 8B + tools (few-shot) & \second{41.20} & \second{51.60} & \second{1.86} \\ \midrule
        \rowcr
        \textbf{TimeART} (trained) & \first{57.05} & \first{59.87} & \first{1.33} \\ \bottomrule
    \end{tabular}}
    \caption{Attribution of improvements. All variants use the identical 21-tool framework and differ only in the tool-use policy. Higher accuracy and lower MAE indicate better performance. \first{Red}: the best, \second{Blue}: the second best.}
    \label{tab:attribution}
\end{table}

\subsection{Comparison with a stronger agentic backbone}
\label{sec:newer backbone}

Since general-purpose LLMs improve rapidly in agentic ability, one may ask whether domain-specific training remains necessary once a newer backbone is used. We therefore evaluate Qwen3.5 9B, a more recent model with notably stronger tool-use pretraining, inside the identical 21-tool framework. As shown in Table~\ref{tab:newer backbone}, the newer backbone does improve over Qwen3 8B without any training, gaining 5.60 points on TimeMQA under few-shot prompting, which confirms that general agentic pretraining transfers to time series tool use to some extent. Nevertheless, TimeART still outperforms it by 10.25 points on TimeMQA and 4.43 points on MTBench reasoning while using the older and smaller Qwen3 8B backbone. Knowing which time series tool fits which analytical situation is thus largely complementary to general instruction tuning, and is precisely what our training supplies. We expect this margin to narrow as backbones keep improving, and we regard tool-augmented training as a cheap way of injecting domain priors rather than as a permanent advantage.

\begin{table}[t]
    \centering
    \resizebox{\linewidth}{!}{
    \begin{tabular}{l|c|c}
    \toprule
        \rowcb
        Model & \makecell{TimeMQA \\ Acc} & \makecell{MTBench \\ Reason. Acc} \\ \midrule
        \rowcg
        Qwen3 8B + tools (few-shot) & 41.20 & 51.60 \\ \midrule
        Qwen3.5 9B + tools (zero-shot) & 43.50 & 53.24 \\ \midrule
        \rowcg
        Qwen3.5 9B + tools (few-shot) & \second{46.80} & \second{55.44} \\ \midrule
        \rowcr
        \textbf{TimeART} (Qwen3 8B, trained) & \first{57.05} & \first{59.87} \\ \bottomrule
    \end{tabular}}
    \caption{A newer backbone with stronger agentic pretraining narrows but does not close the gap to a trained model. All variants share the 21-tool framework. \first{Red}: the best, \second{Blue}: the second best.}
    \label{tab:newer backbone}
\end{table}

\subsection{Effect of the number of available tools}
\label{sec:tool count}

With 21 tools, the action space is large and the tool descriptions consume a substantial share of the context window, which may by itself handicap untrained baselines. We therefore vary the number of tools exposed to the model, following the taxonomy of Table~\ref{tab:tool_categorization}. The 5-tool \textit{core} set keeps one representative tool per major capability, namely \texttt{series\_info}, \texttt{summary\_stats}, \texttt{trend\_classifier}, \texttt{forecaster}, and \texttt{anomaly\_detection}; the 10- and 15-tool sets progressively add the remaining pattern detectors, correlation analyzers, and numerical operators; and the 21-tool set is the full toolset.

Table~\ref{tab:tool count} shows opposite trends for the two models. The untrained Qwen3 8B degrades as tools are added, dropping from 43.48 with the core set to 40.31 with the full set, which is consistent with context overload and with selection difficulty in a large action space. TimeART instead improves monotonically from 48.50 to 57.05, so that the gap widens from 5.02 to 16.74 points. This directly supports the purpose of Stage 1: by internalizing what each tool can and cannot do, the model treats additional tools as additional analytical capability instead of additional confusion.

\begin{table}[t]
    \centering
    \resizebox{0.9\linewidth}{!}{
    \begin{tabular}{c|c|c|c}
    \toprule
        \rowcb
        \# Tools & \makecell{Qwen3 8B \\ (untrained)} & \textbf{TimeART} & Gap \\ \midrule
        \rowcg
        5 (core) & \first{43.48} & 48.50 & +5.02 \\ \midrule
        10 & 42.08 & 53.70 & +11.62 \\ \midrule
        \rowcg
        15 & 39.78 & 55.20 & +15.42 \\ \midrule
        21 (full) & 40.31 & \first{57.05} & +16.74 \\ \bottomrule
    \end{tabular}}
    \caption{TimeMQA accuracy as a function of the number of available tools. More tools hurt the untrained model but help the trained one. \first{Red}: the best per column.}
    \label{tab:tool count}
\end{table}

\subsection{Tool-use quality beyond final-answer accuracy}
\label{sec:tool quality}

Final-answer accuracy does not reveal whether a model invokes tools for the right reasons. We therefore evaluate tool-use quality at the level of individual invocations. We sample 200 evaluation trajectories, stratified over the four TimeMQA categories and over the tool taxonomy of Table~\ref{tab:tool_categorization}, and two annotators independently label every tool invocation along three criteria: \textit{tool selection}, whether the invoked tool is appropriate for the current reasoning state; \textit{parameter configuration}, whether the arguments are set correctly for the question; and \textit{output interpretation}, whether the tool output is read and used correctly in the subsequent thought. Annotators see the trajectory without the gold answer, so as to avoid outcome bias.

Table~\ref{tab:tool quality} compares TimeART with the few-shot prompted Qwen3 8B under the same framework. TimeART is better on all three criteria by more than 20 points, with substantial inter-annotator agreement in both settings. Two further observations indicate that the accuracy gain reflects genuine tool use rather than lucky shortcuts. Among TimeART's correct final answers, 91.2\% also have fully correct intermediate tool use. Among its incorrect answers, 63\% are attributable to the bounded accuracy of the tools themselves, for instance forecasting error on highly volatile series, whereas only 37\% stem from wrong tool selection or parameter errors. All of these numbers come from the same manually verified annotation pass, and a controlled simulation with synthetically known-correct tool sequences would be a natural automatic complement in future work.

\begin{table}[t]
    \centering
    \resizebox{\linewidth}{!}{
    \begin{tabular}{l|c|c}
    \toprule
        \rowcb
        Criterion & \makecell{Qwen3 8B + tools \\ (few-shot)} & \textbf{TimeART} \\ \midrule
        \rowcg
        Correct tool selection & 62.3\% & \first{84.7\%} \\ \midrule
        Correct parameter configuration & 54.8\% & \first{79.2\%} \\ \midrule
        \rowcg
        Correct output interpretation & 58.1\% & \first{82.5\%} \\ \midrule
        Inter-annotator agreement ($\kappa$) & 0.71 & 0.76 \\ \bottomrule
    \end{tabular}}
    \caption{Human evaluation of tool-use quality on 200 sampled evaluation trajectories, judged per tool invocation by two independent annotators. \first{Red}: the best.}
    \label{tab:tool quality}
\end{table}

\subsection{Stability across training runs}
\label{sec:seed variance}

Greedy decoding makes a single evaluation reproducible but says nothing about the variance introduced by fine-tuning, trajectory sampling, or judge-based filtering. We therefore retrain TimeART from scratch with three independent random seeds, keeping the corpus, the hyper-parameters of Table~\ref{tab:lora_hyperparams}, and the evaluation protocol fixed. As Table~\ref{tab:seed variance} shows, the run-to-run standard deviation stays below 0.7 accuracy points, whereas the margin over the strongest untrained variant of Table~\ref{tab:attribution} is 15.85 points on TimeMQA, i.e., more than twenty standard deviations. The conclusions drawn in the main text are therefore not an artifact of a single fortunate run.

\begin{table}[t]
    \centering
    \resizebox{\linewidth}{!}{
    \begin{tabular}{l|c|c|c|c}
    \toprule
        \rowcb
        Metric & Seed 1 & Seed 2 & Seed 3 & Mean $\pm$ Std \\ \midrule
        \rowcg
        TimeMQA Acc & 57.05 & 56.35 & 57.75 & 57.05 $\pm$ 0.70 \\ \midrule
        MTBench Reasoning Acc & 59.87 & 58.97 & 60.27 & 59.70 $\pm$ 0.67 \\ \midrule
        \rowcg
        MTBench Forecast MAE & 1.33 & 1.38 & 1.30 & 1.34 $\pm$ 0.04 \\ \bottomrule
    \end{tabular}}
    \caption{TimeART retrained with three independent seeds. Aggregates follow the definitions of Appendix~\ref{sec:attribution}.}
    \label{tab:seed variance}
\end{table}

\subsection{More analyses on rationale $C$ in Stage 3}

Stage 3 is designed to extract the latent decision principles behind tool selection and convert them into explicit rationales. Although Stages 1--2 already expose the model to tool capability boundaries and expert decision preferences, such knowledge may remain implicit in the model parameters and may not be reliably activated during inference. Therefore, we use the Stage-1--2 trained model to generate rationales $\mathrm{C}$ through self-questioning, where the model is prompted to explain why a particular tool is appropriate for a given reasoning state. These rationales serve as intermediate supervision signals that bridge tool-use behavior and the underlying selection criteria.

The key motivation is that the Stage-1--2 trained model has already acquired empirical experience about when different tools are useful, and can therefore produce more meaningful rationales than an untrained model. However, directly generating rationales in Stage 3 does not by itself guarantee that the model will spontaneously rely on them during future tool selection. To address this, Stage 4 further distills these rationales into the model's output preference, encouraging the model to internalize $\mathrm{C}$ and use it implicitly when deciding which tool to invoke. In this sense, Stages 3--4 form a self-distillation process: Stage 3 externalizes the model's learned tool-selection knowledge into textual rationales, while Stage 4 aligns the model to reproduce tool choices conditioned on such rationales.

We further validate the necessity of using a sufficiently trained model to generate $\mathrm{C}$. Specifically, we compare rationales generated by an untrained model, a Stage-1-only model, and the Stage-1--2 trained model, followed by the same Stage-4 training. The results show that rationales from an untrained model consistently degrade performance, while rationales from a Stage-1-only model do not lead to stable improvements. In contrast, rationales generated after Stages 1--2 provide more reliable supervision, supporting the design choice that rationale extraction should be performed only after the model has learned both tool capability boundaries and expert decision preferences.

\begin{figure*}[t]
  \centering
  \begin{minipage}[b]{0.23\linewidth}
    \centering
    \subfloat[Understanding.]
    {\includegraphics[width=\linewidth]{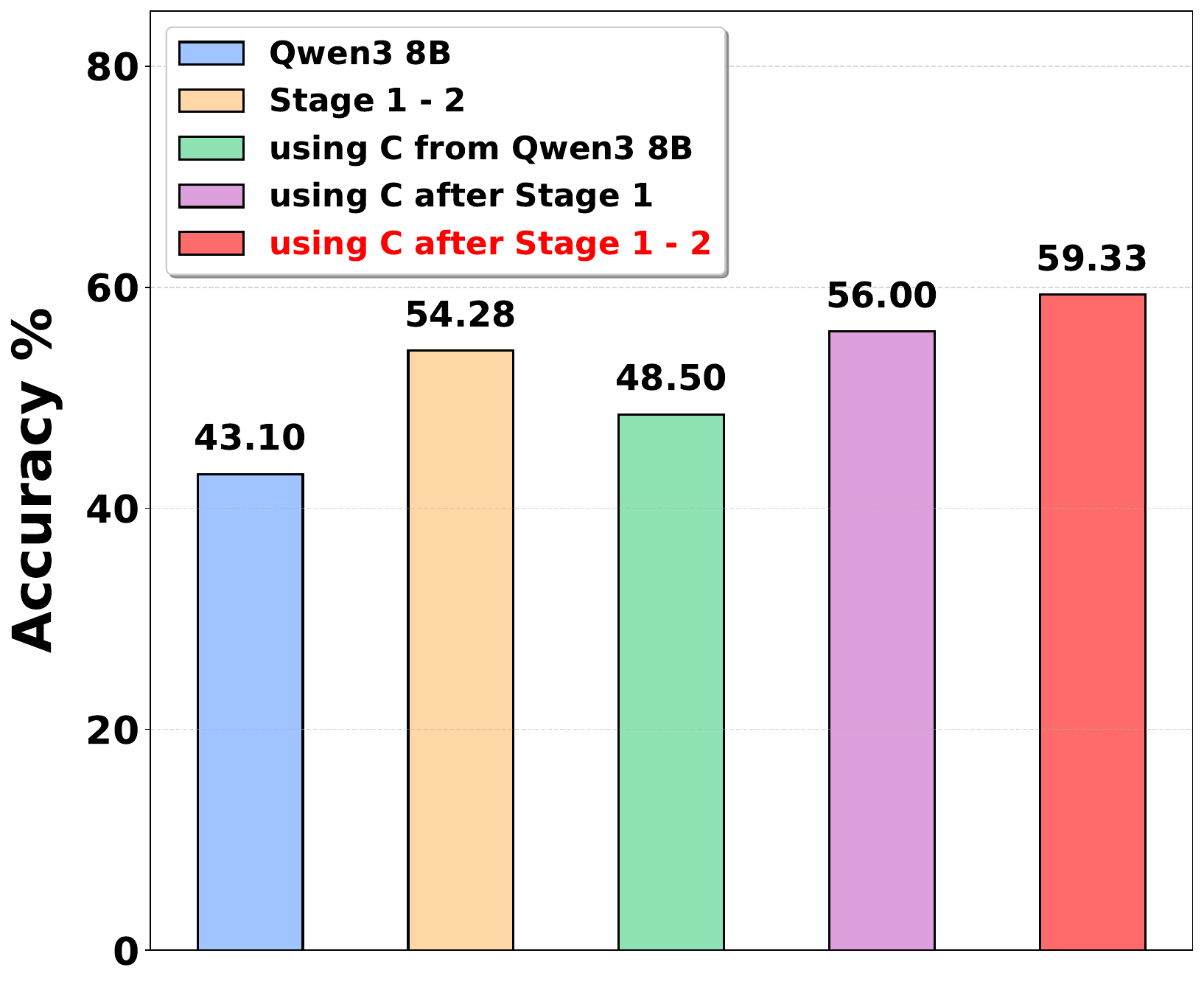}}
  \end{minipage}
  \hfill
  \begin{minipage}[b]{0.23\linewidth}
    \centering
    \subfloat[Perception.]
    {\includegraphics[width=\linewidth]{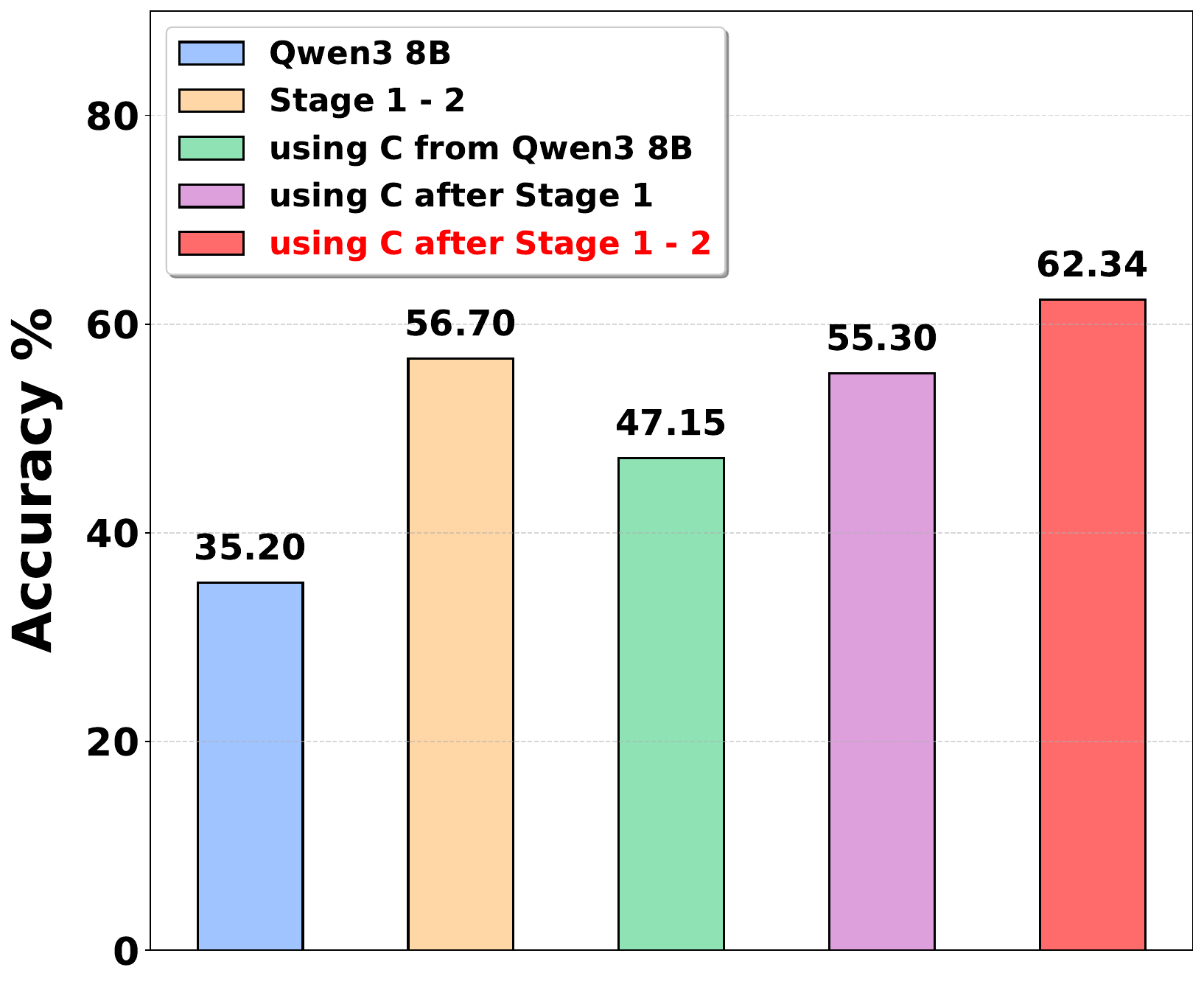}}
  \end{minipage}
  \hfill
  \begin{minipage}[b]{0.23\linewidth}
    \centering
    \subfloat[Reasoning.]
    {\includegraphics[width=\linewidth]{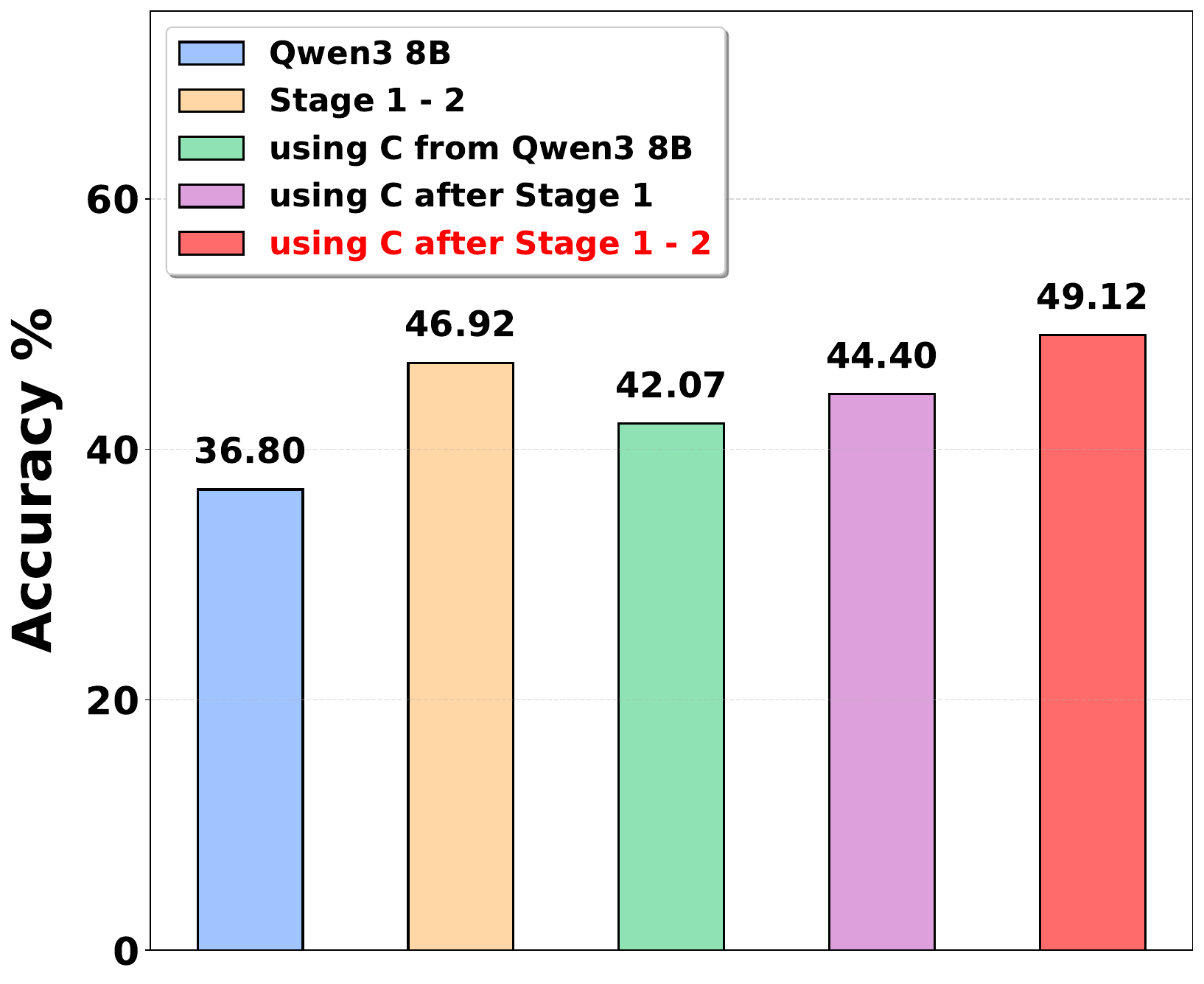}}
  \end{minipage}
  \hfill
  \begin{minipage}[b]{0.23\linewidth}
    \centering
    \subfloat[Estimation.]
    {\includegraphics[width=\linewidth]{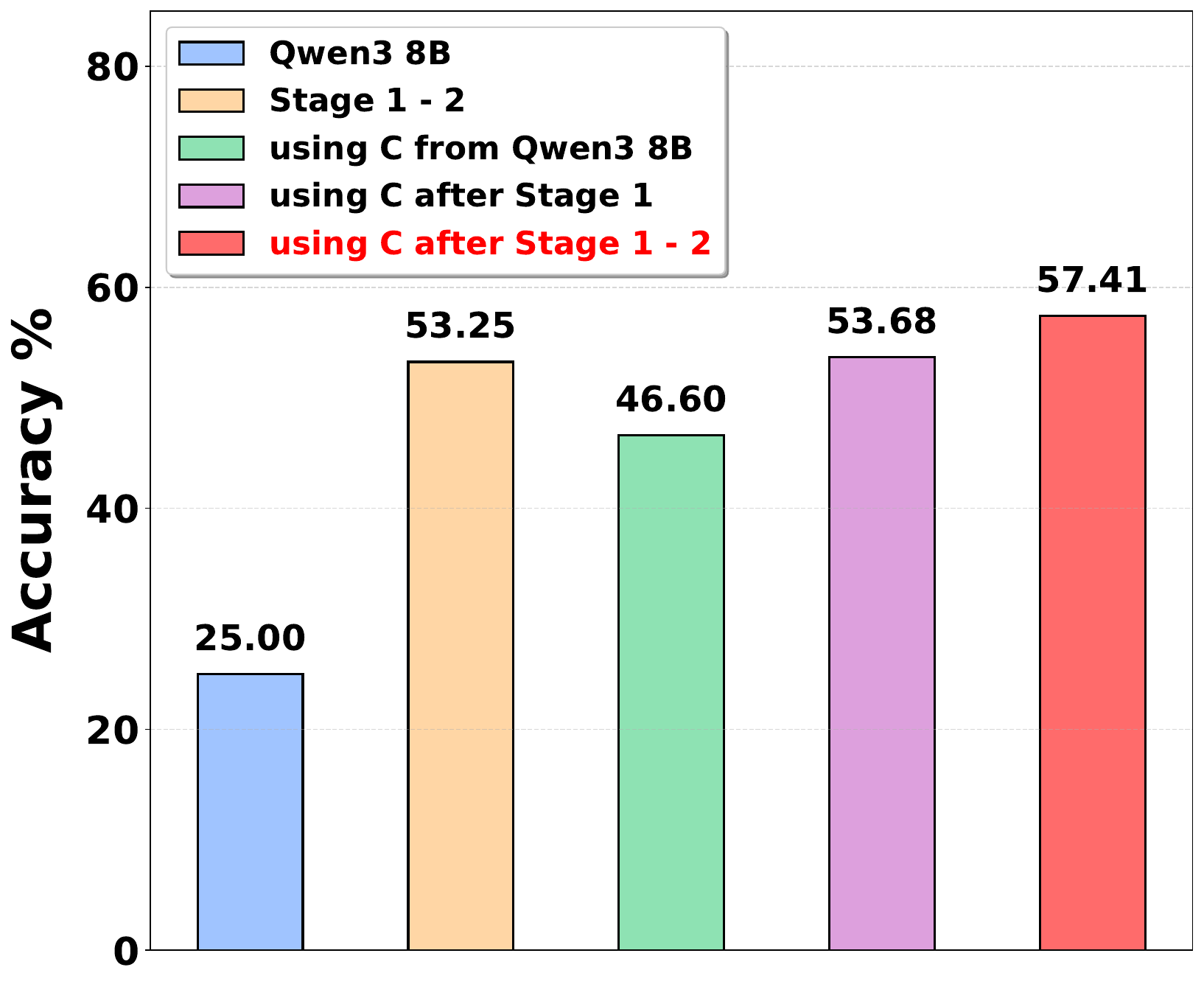}}
  \end{minipage}
  \caption{Analyses on rationale $\mathrm{C}$.}
  \label{fig: analyses on rationale}
\end{figure*}

\section{Implementation details}
\label{app: settings}

\begin{figure}[t]
    \centering
\includegraphics[width=\linewidth]{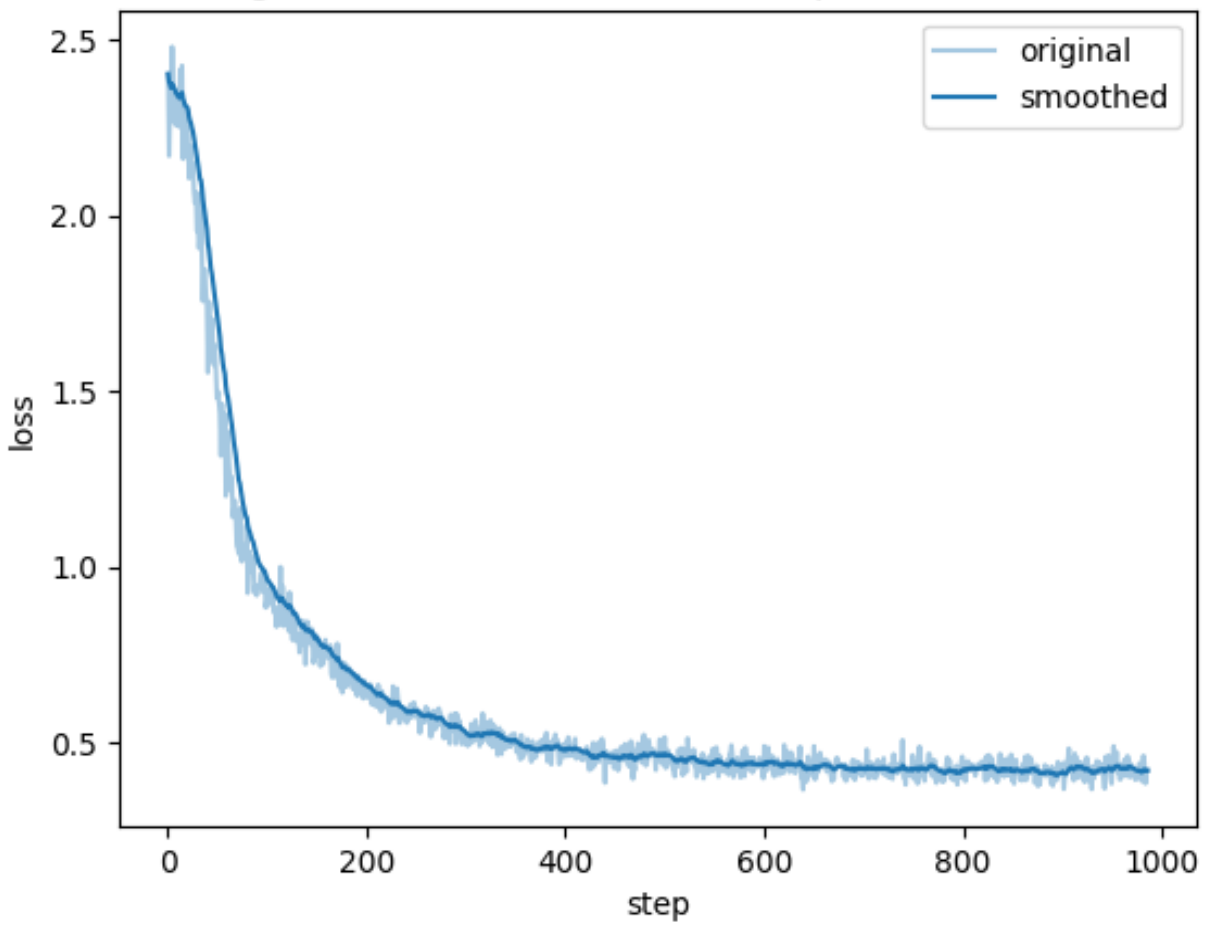}
    \caption{The training curve.}
\label{fig: training curve}
\end{figure}

To ensure the robustness of tool calling, we implement TimeART based on Langchain, the api of all 21 tools are introduced in Appendix~\ref{app: tool}. Since our proposed training strategy does not involve reinforcement learning (RL), all stages can be organized as supervised fine-tuning (SFT). We adopt the LLaMA-Factory to achieve all the training stages, utilizing DeepSpeed ZeRO-3 for efficient training. The fine-tuning is performed in BF16 precision with FlashAttention-2 enabled to accelerate attention operations. We utilize LoRA, please refer to Table~\ref{tab:lora_hyperparams} for detailed configurations. The Training Curve is shown in Figure~\ref{fig: training curve}. When evaluating the open-source LLMs (including the baselines and TimeART), we set $\texttt{do\_sample=False}$ to ensure reproducibility. This forces LLMs to use Greedy Decoding, strictly selecting the token with the maximum logit at each step, which guarantees that the results are reproducible on the same device. 

\begin{table}[h]
    \centering
    \resizebox{0.6\linewidth}{!}{
    \footnotesize
    \renewcommand{\arraystretch}{1}{
    \begin{tabular}{l l}
        \toprule
        \textbf{Hyperparameter} & \textbf{Assignment} \\
        \midrule
        Base model & Qwen3 8B \\
        Computing infrastructure & 8*3090-24GB GPU \\
        Epochs & 5 \\
        Warm-up ratio & 0.02 \\
        Batch size per device & 4 \\
        Gradient accumulation steps & 16 \\
        Learning rate & 1e-5 \\
        Embedding learning rate & 1e-5 \\
        Optimizer & AdamW \\
        Learning rate scheduler & Cosine \\ \midrule
        LoRA rank (r) & 8 \\
        LoRA alpha & 16 \\
        LoRA dropout & 0.05 \\
        LoRA target modules & q\_proj, k\_proj, v\_proj, o\_proj, \\
        & gate\_proj, up\_proj, down\_proj \\ \midrule
        Training time & \textasciitilde2 Day \\ 
        \bottomrule
    \end{tabular}}}
    \caption{Hyper-parameters and training time of fine-tuning the Qwen3 8B based on TimeToolBench.}
    \label{tab:lora_hyperparams}
\end{table}

\section{Dataset Details}
\label{app: dataset}

\textbf{Benchmarks}

MTBench is a large-scale multimodal benchmark designed to evaluate the capacity of Large Language Models (LLMs) to integrate and reason across paired time-series and textual data, addressing a critical gap in measuring cross-modal understanding. 

Spanning the finance and weather domains, the benchmark aligns numerical data—such as historical stock prices and weather records—with corresponding narratives like financial news and storm reports to simulate complex real-world analysis. Unlike traditional benchmarks that focus on isolated modalities or simple prediction, MTBench challenges models with a suite of reasoning-intensive tasks, including time-series forecasting, technical indicator prediction, semantic trend analysis, and news-driven question answering. 

Time-MQA is a unified framework designed to enable multi-task question answering for time series through context enhancement, moving beyond the traditional focus on single numerical tasks. At its core lies the TSQA dataset, a large-scale collection of over 200,000 question-answer pairs spanning 12 diverse domains, including healthcare, finance, energy, and traffic. Time-MQA integrates five major categories of tasks—forecasting, imputation, anomaly detection, classification, and open-ended reasoning—into a single question-answering paradigm, allowing users to interact with data via natural language queries supported by textual context. 

In this work, we evaluate on most tasks in MTBench, and choose the subset of TimeMQA, classify them into four categories. The purpose is to ensure that the benchmark datasets do not overlap with TimeToolBench and to evaluate as many types of tasks as possible.

\textbf{TimeToolBench}

As stated in Section~\ref{sec: timetoolbench}, we employ three different models to evaluate the quality of the reasoning chains: DeepSeek-V3.2, Claude Opus 4.5, and Gemini 3 Pro. By using models from three different vendors, we aim to diversify the evaluation criteria and strive for a more rigorous consensus.
In practice, if a trajectory generated by GPT-4o yields the correct final answer but the reasoning logic is rejected, we prompt GPT-4o by indicating that the result is correct but the process is flawed, asking it to refine its reasoning. If it still fails to pass the evaluation after a limited number of attempts ($\leq 10$), only then do we completely discard this data point. Consequently, the overall discard rate remains relatively low ( < 10\%).

Because the trajectories are generated with knowledge of the gold answer, a legitimate concern is that some of them rationalize the answer post hoc instead of genuinely solving the problem. We therefore conduct a human validation on 100 trajectories sampled uniformly from the corpus. Annotators do not see the answer-generation prompt and judge whether the intermediate reasoning is valid and, in particular, whether the same tool would plausibly be chosen without knowing the answer. 78\% of the trajectories are rated as fully valid with non-trivial tool use, 14\% reach the correct answer but contain at least one suboptimal step, and 8\% are clear post-hoc rationalizations, with an inter-annotator agreement of $\kappa = 0.73$. On the same sample, our automatic judge pipeline agrees with the human verdict 91\% of the time. Some degree of answer conditioning is inherent to distillation-based trajectory generation; the high validation rate together with the held-out results of Table~\ref{tab:attribution} indicates that the corpus teaches transferable tool-use principles rather than answer-specific explanations.

\begin{table}[htbp]
\centering
\renewcommand{\arraystretch}{1.2}
\resizebox{\linewidth}{!}{%
\begin{tabular}{@{}lll@{}}
\toprule
\textbf{Evaluation set} & \textbf{Trajectories in TimeToolBench} & \textbf{Exclusion method} \\
\midrule
MTBench (all tasks) & 22k & entirely removed \\
TimeMQA (four subsets) & 2.6k & held out by domain \\
\midrule
\textit{Remaining training set} & \textit{75.4k} & \textit{non-overlapping domains} \\
\bottomrule
\end{tabular}}
\caption{Separation between TimeToolBench and the evaluation benchmarks. No raw series, question template, or answer label from an evaluation set appears in training.}
\label{tab:split}
\end{table}

\textbf{Training Data}

TimeToolBench covers trajectories from TimeSeriesExam, TimeSeriesGym, MTBench, and TimeMQA. During the training of TimeART, we remove all MTBench trajectories (22k). As for TimeMQA, which spans multiple domains, we hold out datasets from specific domains (four subsets, 2.6k) purely for evaluation, ensuring no corresponding trajectory data leaks during training. In summary, we utilize 75.4k out of the 100k trajectories in TimeToolBench for training. Table~\ref{tab:split} summarizes this separation. The domains of these training trajectories have no overlap with the downstream evaluation datasets, thus constituting a rigorous Out-of-Distribution (OOD) evaluation. The raw training data in TimeToolBench is collected and organized in ReAct-style format. The characteristic of ReAct-style data is that the structure of reasoning is clear, and the steps always follow the established order of Thought-Action-Observation. We first show a simple example as follows:

\begin{tcolorbox}[
  breakable,
  colback=white,
  colframe=black!50,
  boxrule=0.3mm,
  width=\columnwidth,
  rounded corners,
  title=A ReAct-style Example,
  fonttitle=\bfseries,
  left=2mm, right=2mm, top=2mm, bottom=2mm
]
\textbf{Query}:\\
Here is the time series. Please analyze its pattern.\\

\textbf{Thought}:\\
The most important patterns are trend and seasonality. I will detect them separately.\\

\textbf{Action}:\\
tool: [trend\_classifier]\\

\textbf{Observation}:\\
tool: [trend\_classifier]\\
output: The time series shows up trend ...\\

\textbf{Thought}:\\
I will detect the seasonality next step. \\

\textbf{Action}:\\
tool: [seasonality\_detector]\\

\textbf{Observation}:\\
tool: [seasonality\_detector]\\
The most clear period seems to be 24 ... \\

\textbf{Final Answer}:\\
The patterns are ...
\end{tcolorbox}

\begin{tcolorbox}[
  breakable,
  colback=white,
  colframe=black!50,
  boxrule=0.3mm,
  width=\columnwidth,
  rounded corners,
  title=sharegpt format,
  fonttitle=\bfseries,
  left=2mm, right=2mm, top=2mm, bottom=2mm
]
\textbf{from human}:\\
Here is the time series. Please analyze its pattern.\\

\textbf{from gpt}:\\
The most important patterns are trend and seasonality. I will detect them separately.\\

\textbf{from human}:\\
(continue)\\

\textbf{from function\_call}:\\
tool: [trend\_classifier]\\

\textbf{from observation}:\\
tool: [trend\_classifier]\\
output: The time series shows up trend ...\\

\textbf{from gpt}:\\
I will detect the seasonality next step. \\

\textbf{from human}:\\
(continue)\\

\textbf{from function\_call}:\\
tool: [seasonality\_detector]\\

\textbf{from observation}:\\
tool: [seasonality\_detector]\\
The most clear period seems to be 24 ... \\

\textbf{from gpt}:\\
The patterns are ...
\end{tcolorbox}

During the training process, we hope the overall logic of the context will be preserved. However, this format does not seem suitable for SFT LLMs using the question-answering paradigm. 

To handle this problem, we choose to use the most-liked sharegpt format in LLaMA-Factory. And the above-mentioned example can be organized as multi-turn conversations. Note that the human and observation should appear in odd positions, while gpt and function should appear in even positions. The gpt and function will be learned by the model.

\begin{table}[htbp]
\centering
\renewcommand{\arraystretch}{1.3}
\resizebox{\linewidth}{!}{%
\begin{tabular}{@{}ll@{}}
\toprule
\textbf{Category} & \textbf{Associated Tools} \\
\midrule
Forecasting       & \texttt{forecaster} \\
Anomaly Detection & \texttt{anomaly\_detection}, \texttt{spike\_detector} \\
Classification    & \texttt{trend\_classifier}, \texttt{noise\_profile}, \texttt{stationarity\_test} \\
Seasonality       & \texttt{seasonality\_detector}, \texttt{autocorr} \\
Trend             & \texttt{trend\_classifier}, \texttt{rolling\_stat}, \texttt{summary\_stats} \\
Stationarity      & \texttt{stationarity\_test} \\
Transition        & \texttt{change\_point\_detector} \\
Shifting          & \texttt{cross\_correlation}, \texttt{dtw\_distance} \\
Correlation       & \texttt{channel\_correlation}, \texttt{cross\_correlation}, \texttt{autocorr}, \\
                  & \texttt{shape\_similarity}, \texttt{granger\_causality} \\
\midrule
\textit{Other (Basic Stats)} & \textit{\texttt{series\_info}, \texttt{datapoint\_value}, \texttt{return\_calc},} \\
                             & \textit{\texttt{quantile\_value}, \texttt{volatility}} \\
\bottomrule
\end{tabular}}
\caption{Categorization of TimeART Tools based on Analytical Tasks}
\label{tab:tool_categorization}
\end{table}

\section{Details of Tools}
\label{app: tool}

We design the toolset based on prior knowledge from the time-series analysis domain. Referring to perspectives proposed by well-recognized benchmarks such as TFB~\citep{qiu2024tfb} and BLAST~\citep{DBLP:conf/kdd/Shao00YFQ0D0C25}, the difficulty of time-series analysis primarily relates to the inherent properties of the data, among which Seasonality, Trend, Stationarity, Transition, Shifting, and Correlation dominate. On the other hand, common tasks of time-series analysis include forecasting, anomaly detection, and classification. Therefore, the "atomic design" we mention aims to equip the tools with the ability to perceive these properties and resolve common tasks.
We also provide a taxonomy table classifying the 21 tools according to the aforementioned categories in Table~\ref{tab:tool_categorization}. As shown, these tools balance coverage of the main elements mentioned above.

\textbf{Numerical Operators}

\noindent-- \texttt{series\_info}. (). Retrieves basic metadata of the time series, including sequence length (T), number of channels (C), and missing value statistics.

\noindent-- \texttt{datapoint\_value}. (\texttt{index\_or\_timestamp}). Returns the specific values of all channels at a given time index or timestamp.

\noindent-- \texttt{summary\_stats}. (\texttt{start}, \texttt{end}, \texttt{stat}). Calculates a specific statistic (mean, sum, max, min, std) for all channels over a defined index range [start, end).

\noindent-- \texttt{return\_calc}. (\texttt{t1},\texttt{t2}, \texttt{kind}). Computes the percentage return (``pct'') or absolute difference (``diff'') between two specific time indices.

\noindent-- \texttt{autocorr}. (\texttt{lag}). Computes the autocorrelation coefficient for each channel at a specified time lag to measure self-similarity.

\noindent-- \texttt{rolling\_stat}. (\texttt{stat}, \texttt{window}, \texttt{step}). Computes rolling statistics (mean, sum, max, min, std) using a sliding window across the time series.

\noindent-- \texttt{quantile\_value}. (\texttt{q}). Calculates the empirical value at a specific quantile level (between 0 and 1) for each channel (e.g., q=0.5 for median).

\noindent-- \texttt{volatility}. (\texttt{window}). Computes rolling volatility (calculated as the standard deviation of first differences) over a specified window size.

\textbf{Pattern Detector}

\noindent-- \texttt{trend\_classifier}. (\texttt{window}). Classifies trends in time series as ``up'', ``down'', or ``flat''; supports global analysis or window-based segment analysis.

\noindent-- \texttt{seasonality\_detector}. (\texttt{max\_period}). Detects periodic patterns and returns estimated period with seasonality strength (``strong'' or ``weak'').

\noindent-- \texttt{change\_point\_detector}. (\texttt{penalty\_or\_n\_cp}).  Detects structural breaks (change points) in mean or variance and returns the indices of these changes.

\noindent-- \texttt{noise\_profile}. (\texttt{window}). Labels noise type (e.g., ``white'', ``red'') based on autocorrelation tests; performed globally or over a specific window.

\noindent-- \texttt{stationarity\_test}. (\texttt{test}). Tests stationarity using Augmented Dickey-Fuller or KPSS methods; returns status (``stationary''/``nonstationary'') and test statistics.

\noindent-- \texttt{spike\_detector}. (\texttt{threshold}, \texttt{min\_sep}). Detects and locates spikes or dips in the series based on amplitude threshold and minimum separation.

\textbf{Correlation Analyzer}

\noindent-- \texttt{channel\_correlation}. (\texttt{channel\_1}, \texttt{channel\_2}, \texttt{lag}, \texttt{method}). Calculates correlation (``Pearson''/``Spearman'') between two channels with an optional time lag.

\noindent-- \texttt{cross\_correlation}. (\texttt{channel\_1}, \texttt{channel\_2}, \texttt{max\_lag}). Computes cross-correlation across multiple lags to find the optimal time alignment between two channels.

\noindent-- \texttt{dtw\_distance}. (\texttt{channel\_1}, \texttt{channel\_2}, \texttt{distance\_metric}). Measures similarity between two channels using Dynamic Time Warping (DTW); returns distance where lower values indicate greater similarity.

\noindent-- \texttt{shape\_similarity}. (\texttt{channel\_1}, \texttt{channel\_2}, \texttt{norm}). Measures shape similarity between two channels using normalized correlation, invariant to amplitude scaling.

\noindent-- \texttt{granger\_causality}. (\texttt{cause\_channel}, \texttt{effect\_channel}, \texttt{max\_lag}). Tests if one channel statistically predicts another (Granger causality) within a specified maximum lag.

\textbf{Forecasting and Anomaly Detection}

\noindent-- \texttt{anomaly\_detection}. (\texttt{anomaly\_threshold}). Detects anomalies in multivariate time series using the state-of-the-art zero-shot detector DADA~\citep{shentu2024towards} based on reconstruction error (MSE); selects the most significant anomalies by interpreting the threshold as a top percentage (if 0-1) or a specific count (if $\ge$ 1).

\noindent-- \texttt{forecaster}. (\texttt{forecast\_horizon}). Generates forecasts for multivariate time series using the state-of-the-art zero-shot forecaster LightGTS~\citep{wang2025lightgts}; returns predicted values for all channels.

\textbf{Extensibility of the Toolset}

Because every tool is described in the external constraint $\mathcal{E}$ and its capability boundary is learned independently in Stage 1, the toolset can be modified without retraining TimeART from scratch. Removing a tool only requires deleting its description from $\mathcal{E}$; the model then stops calling it, since nothing in the learned policy binds a reasoning state to a tool that is no longer offered. In a controlled check, removing three randomly chosen tools at test time changes accuracy on tasks that need only the remaining tools by less than 0.5\%, indicating that the learned policy is separable across tools rather than tied to the exact 21-tool configuration. Adding or replacing a tool is lightweight but not entirely free: it involves an additional Stage 1 pass over a small set of trajectories for the new tool, between 500 and 1000 samples, so that its capability boundary is learned, while the Stage 2--4 knowledge for the other tools is preserved. Registering a new tool with no gradient update at all is an appealing extension that we leave to future work.

\section{Further Related Work}
\label{sec:further related}
Besides Time Series Reasoning, conventional research focuses on Time Series Analysis, which holds a position of paramount importance across a diverse range of fields, including the economy~\citep{qiu2025easytime,qiu2025comprehensive,liu2025calf}, transportation~\citep{wu2024fully,AutoCTS++,qiu2025duet}, health \citep{lu2023tf,lu2024mace,lu2024robust}, weather \citep{li2025set,yang2024wcdt,zhou2025reagent}, and energy \citep{sun2025hierarchical}. It encompasses a multitude of critical tasks, such as forecasting~\citep{wu2025k2vae,wu2025srsnet}, anomaly detection~\citep{wang2025unitmge,qiu2025tab,wu2024catch}, classification~\citep{liu2023itransformer}, imputation~\citep{wu2022timesnet}, and others~\citep{qiu2025dag,qiu2025DBLoss,liu2026apn}. 

Researchers also make efforts to construct time series foundation models to support zero-shot analytical tasks like forecasting~\citep{liu2025sundial,wang2025lightgts,woo2024moirai}, anomaly detection~\citep{units,shentu2024towards}, and classification~\citep{AimTS}. In this work, we adopt them as strong out-of-the-box tools, endowing TimeART the numerical capability.

Among the LLM-based approaches to conventional analytical tasks discussed in Section~\ref{sec: related works}, TimeLLM~\citep{timellm}, TimeVLM~\citep{timevlm}, ChatTime~\citep{wang2025chattime}, UniTime~\citep{liu2024unitime}, and CALF~\citep{liu2025calf} demonstrate certain performance. With the rise of exploration in TSQA, several benchmarks are proposed to fairly evaluate TSRMs, among which MTBench~\citep{chen2025mtbench}, TimeSeriesExam~\citep{cai2024timeseriesexam}, and TimeSeriesGym~\citep{cai2025timeseriesgym} gain attention. Time series agents have also been built for conventional tasks such as representation learning~\citep{zhou2025merit}, classification~\citep{sui2025training}, and AutoML for forecasting~\citep{zhao2025timeseriesscientist,wang2025aries}, so that they can replace repetitive labor to some extent.

\section{Extended Discussion of Limitations}
\label{app:limitations}

Despite the strong performance of TimeART on diverse time series reasoning and forecasting tasks, this work still presents limitations, such as the framework may face efficiency and context retention challenges when processing extremely long or high-dimensional multivariate time series, the TimeToolBench corpus built from public datasets and GPT-4o trajectories still lacks sufficient coverage of edge cases in some industrial and medical scenarios.

Our experiments also point to several directions along which the present evidence can be strengthened further. Table~\ref{tab:attribution} keeps the toolset fixed across all variants and thereby separates the training strategy from tool access, and the per-stage ablation of Figure~\ref{fig: ablations on training} gives a first picture of how much each stage contributes; a study that additionally varies the amount of training data would sharpen this attribution. Since TimeToolBench is generated with knowledge of the gold answers, a small fraction of the trajectories contains suboptimal intermediate steps, and refining the generation prompts or adding a verification stage is a natural way to raise the proportion of fully clean demonstrations. Our filtering and part of our evaluation use LLM judges, which agree closely with human annotators on the samples we checked, and pairing them with lightweight human spot checks would make the pipeline more robust still. The evaluation benchmarks, although disjoint from the training trajectories at the level of raw series, templates, and labels, come from related benchmark families, so extending the evaluation to further domains such as industrial or medical monitoring would broaden the evidence for generality. Finally, the process-level evaluation of Appendix~\ref{sec:tool quality} is based on human annotation of a sample, which a controlled simulation with known-correct tool sequences could complement with a fully automatic measurement.

\section{Impact Statement}
\label{app:impact}

This paper automates time series analysis, lowers technical barriers for non-experts, supports applications in finance, healthcare, weather forecasting, and smart grids, and provides open resources to boost tool-augmented LLM research, while potential risks such as erroneous decisions from over-reliance on tools and dataset biases are mitigated by requiring human supervision in high-stakes scenarios. And we only use LLMs for grammatical polishing during writing, with all research content being original and ethically compliant.

\section{Prompt Templates}
\label{app: prompt}

\begin{tcolorbox}[
  breakable,
  colback=white,
  colframe=black!50,
  boxrule=0.3mm,
  width=\columnwidth,
  rounded corners,
  title=Prompt 1: Making the self-reflections in Stage 3,
  fonttitle=\bfseries,
  left=2mm, right=2mm, top=2mm, bottom=2mm
]

You will be presented with a situation where you need to choose between multiple possible actions.
Your task is to analyze the situation and provide reasoning about why we decide to take the expert action.

\begin{itemize}[leftmargin=1em,labelsep=0.6em,itemsep=0.4ex,topsep=0.5ex]
  \item \textbf{Situation Description ($\mathcal{S}_k$):} \{\texttt{Situation Description}\}
  \item \textbf{Expert Action ($\mathrm{A}_k$):} \{\texttt{Expert Action}\}
  \item \textbf{Expected Outcome ($\mathrm{O}_k$):} \{\texttt{Future State of Expert Action}\}
  \item \textbf{Alternative Actions:}
    \begin{enumerate}[leftmargin=1em,labelsep=0.6em,itemsep=0.4ex,topsep=0.5ex]
      \item Action $\mathrm{A}_k^1$: \{\texttt{Alt Action 1}\}, resulting state $\mathrm{O}_k^1$: \{\texttt{State 1}\}
      \item Action $\mathrm{A}_k^2$: \{\texttt{Alt Action 2}\}, resulting state $\mathrm{O}_k^2$: \{\texttt{State 2}\}
      \item \dots
    \end{enumerate}
\end{itemize}

Provide a detailed self-reflection as an \emph{internal monologue} that demonstrates your reasoning process for the current situation.
Your monologue should:

\begin{enumerate}[leftmargin=2em,labelsep=0.6em,itemsep=0.4ex,topsep=0.5ex]
  \item Analyze the situation and the goal.
  \item Compare the possible actions, explaining why each may be less optimal.
  \item Justify why the expert action is most suitable, grounded in the expected outcome.
  \item Highlight any relevant clues, constraints, or consequences from the situation.
\end{enumerate}

\textbf{Guidelines:}
\begin{itemize}[leftmargin=1em,labelsep=0.6em,itemsep=0.4ex,topsep=0.5ex]
  \item Stay strictly within the provided information.
  \item Avoid meta-commentary about being an AI.
  \item Use natural, step-by-step reasoning.
  \item Focus on logical decision-making.
\end{itemize}

\textbf{Output:} Directly write the self-reflection monologue, no extra headings, disclaimers, or external notes.

\end{tcolorbox}

\begin{tcolorbox}[
  breakable,
  colback=white,
  colframe=black!50,
  boxrule=0.3mm,
  width=\columnwidth,
  rounded corners,
  title=Prompt 2: Collecting the TimeToolBench,
  fonttitle=\bfseries,
  left=2mm, right=2mm, top=2mm, bottom=2mm
]

You are an intelligent Time Series Reasoner capable of performing time series analysis by invoking appropriate tools step by step. \\
\begin{enumerate}[leftmargin=1em,labelsep=0.6em,itemsep=0.4ex,topsep=0.5ex]
  \item Given time series data, a question, and the known answer, reconstruct the intermediate reasoning steps.
  \item You should understand the time series and use the tools to enhance your confidence. You shouldn't completely rely on the results from tools.
  \item You should call a tool at least once to help answer the question. More tool calls are encouraged.
  \item You should think step by step in ReAct-style, output a structured reasoning trajectory that leads to the final answer.
\end{enumerate}
You have access to the following tools: \{tools\}
The only tools you may use are: \{tool\_names\}. \\
"Begin!"\\

The question is: \{input\}
\end{tcolorbox}

\begin{tcolorbox}[
  breakable,
  colback=white,
  colframe=black!50,
  boxrule=0.3mm,
  width=\columnwidth,
  rounded corners,
  title=Prompt 3: Evaluation,
  fonttitle=\bfseries,
  left=2mm, right=2mm, top=2mm, bottom=2mm
]

You are an intelligent Time Series Reasoner capable of performing time series analysis by invoking appropriate tools step by step.\\
\begin{enumerate}[leftmargin=1em,labelsep=0.6em,itemsep=0.4ex,topsep=0.5ex]
  \item You should first understand the question and analyze whether it is necessary to invoke the tool. If not, you can directly give the final answer.
  \item If tools are needed, you can utilize them to help answer the question. Multiple tool calls are encouraged. 
  \item You should think step by step in ReAct-style, output a structured reasoning trajectory that leads to the final answer.
\end{enumerate}
You have access to the following tools: \{tools\}
The only tools you may use are: \{tool\_names\}. \\
"Begin!"\\

The question is: \{input\}
\end{tcolorbox}

\section{Case Study}
\label{app: timetoolbench}

\begin{tcolorbox}[
  breakable,
  colback=white,
  colframe=black!50,
  boxrule=0.3mm,
  width=\columnwidth,
  rounded corners,
  title=Sample 1: Anomaly Detection,
  fonttitle=\bfseries,
  left=2mm, right=2mm, top=2mm, bottom=2mm
]

\textbf{Query}:\\
Which value is likely an outlier or anomaly in the dataset [0.06, 0.11, 0.13, 0.14, 0.12, 0.13, 0.15, 0.14, 0.15, \textcolor{red}{0.51}, 0.16, 0.13, 0.12, 0.14, 0.11, 0.10, 0.09, 0.08, 0.07, 0.06]? \\ A) 0.51 B) 0.15 C) 0.14 D) 0.09, \\

\textbf{Thought}:\\
To identify the outlier or anomaly in the dataset, I can use the [anomaly\_detector] tool, which detects anomalies based on reconstruction error. This will help confirm which value is most likely an anomaly.\\

\textbf{Action}:\\
tool: [anomaly\_detector], tool\_input: \{"anomaly\_threshold": 1\}\\

\textbf{Observation}: \\
tool: [anomaly\_detector], 
\\
output: anomaly\_scores: [9.589107503416017e-05, 0.001930518657900393, 0.000164090160978958, 0.0014878634829074144, 0.012693900614976883, 0.021373838186264038, 0.022725816816091537, 0.03277913108468056, 0.05729445442557335, \textcolor{red}{0.1406012624502182}, 3.538517034940014e-09, 0.00018555686983745545, 2.8186357667436823e-05, 2.174061464188526e-08, 4.634935066860635e-06, 0.00020597095135599375, 3.8789648897363804e-06, 0.0008871419704519212, 0.0001234006485901773, 0.0007429602555930614] \\

\textbf{Final Answer}: \\
A) 0.51
\end{tcolorbox}

\begin{tcolorbox}[
  breakable,
  colback=white,
  colframe=black!50,
  boxrule=0.3mm,
  width=\columnwidth,
  rounded corners,
  title=Sample 2: Statistical Analysis,
  fonttitle=\bfseries,
  left=2mm, right=2mm, top=2mm, bottom=2mm
]
\textbf{Query}:\\
Calculate and interpret the 3-point moving average of the following sequence [4875.91, 4871.64, 4875.75, 4879.98, 4878.27, 4880.68]. \\

\textbf{Thought}:\\
I need to compute the 3-point moving average for the given sequence. This can be done using the [rolling\_stat] tool with a window size of 3 and the "mean" statistic.\\

\textbf{Action}:\\
tool: [rolling\_stat], tool\_input: \{"stat": "mean", "window": 3, "step": 1\}\\

\textbf{Observation}: \\
tool: [rolling\_stat], 
\\
output: \{"statistic": "mean", "window\_size": 3, "step\_size": 1, "rolling\_results": \{"channel\_0": [\{"window\_start": 0, "window\_end": 3, "mean": 4874.433333333333\}, \{"window\_start": 1, "window\_end": 4, "mean": 4875.79\}, \{"window\_start": 2, "window\_end": 5, "mean": 4878.0\}, \{"window\_start": 3, "window\_end": 6, "mean": 4879.643333333333
\}]\}\}"\} \\

\textbf{Final Answer}: \\
The 3-point moving averages are approximately [4874.43, 4875.79, 4878.00, 4879.64], indicating an upward trend in the sequence.
\end{tcolorbox}

\begin{tcolorbox}[
  breakable,
  colback=white,
  colframe=black!50,
  boxrule=0.3mm,
  width=\columnwidth,
  rounded corners,
  title=Sample 3: Compound task,
  fonttitle=\bfseries,
  left=2mm, right=2mm, top=2mm, bottom=2mm
]
\textbf{Query}:\\
Your task is to analyze the past 14 days's of temperature trend and predict the temperature trend for the next 3 days'. The input time series represents temperature readings from 2021-08-19 00:00:00 to 2021-09-01 23:00:00, with a granularity of hourly. Every 24 temperature readings represent a full day from 00:00 to 23:00. The input temperature readings are: [23.3, 22.567, 21.833, 21.1, 20.733, 20.367, 20.0, 20.0, 20.0, 20.0, 20.0, $\cdots$, 19.8, 19.6, 19.4, 19.233, 19.067, 18.9, 18.9, 18.9, 18.9, 19.067, 19.233, 19.4, 19.233, 19.067, 18.9, 19.067, 19.233, 19.4, 19.4, 19.4]. The following events were reported: Flash Flood, Heavy Rain. These occurred near station USW00094823, approximately 23.0844 km away, between 2021-09-01 00:00:00 and 2021-09-01 13:19:00. $\cdots$ Based on the information you received, predict the temperature trend for the next 3 days. Calculate the mean temperature of the last 24-hour period (i.e., the most recent day in the input) and compare it with the mean temperature of the first predicted day. If the difference is greater or equal than 0.5, classify the trend as 'increasing'. If the difference is less or equal than -0.5, classify the trend as 'decreasing'. Otherwise, classify it as 'stable'.\\

\textbf{Thought}:\\
I need to forecast the next 3 days of temperature and compute the mean of the last day in the input and the first predicted day to determine the trend.\\

\textbf{Action}:\\
tool: [forecaster], tool\_input: \{"forecast\_horizon": 72\} \\

\textbf{Observation}: \\
tool: [forecaster], 
\\
output: forecasts: \{'channel\_0': [23.591698, 22.918169, 22.337996, 21.434757, 21.20016, 21.091335, 20.976793, 20.903154, 20.723076, 20.79474, 20.872782, 20.757236, 20.866882, 21.556917, 22.472641, 23.420639, 24.036705, 24.799648, 25.457306, 25.679733, 25.808376, 26.042328, 25.737324, 25.422695, 23.985361, 23.52016, 22.932299, 22.15012, 21.787104, 21.76454, 21.582138, 21.461386, 21.27549, 21.25565, 21.36626, 21.097694, 21.083847, 21.87988, 22.717966, 23.65874, 24.145344, 24.658247, 25.030067, 25.2354]\} \\

\textbf{Thought}: \\
I need to calculate the mean temperature of the last 24-hour period in the input data and the mean of the first 24 predicted values, then compare them to determine the trend. \\

\textbf{Action}: \\
tool: [summary\_stats], tool\_input: \{"start": 312, "end": 336, "stat": "mean"\} \\

\textbf{Observation}:\\
tool: [summary\_stats],\\
output: \{"range": \{"start": 312, "end": 336\}, "statistic": "mean", "results": \{"channel\_0": 19.3875\}\}"\} \\

\textbf{Thought}: \\
I need to calculate the mean of the first 24 predicted values (first predicted day) to compare with the last 24-hour mean. \\

\textbf{Action}: \\
tool: [summary\_stats], tool\_input: \{"start": 0, "end": 24, "stat": "mean"\} \\

\textbf{Observation}:\\
tool: [summary\_stats],\\
output: \{"range": \{"start": 0, "end": 24\}, "statistic": "mean", "results": \{"channel\_0": 22.871\}\}"\} \\

\textbf{Final Answer}:\\
increasing.
\end{tcolorbox}

\begin{tcolorbox}[
  breakable,
  colback=white,
  colframe=black!50,
  boxrule=0.3mm,
  width=\columnwidth,
  rounded corners,
  title=Sample 4: Trend Prediction,
  fonttitle=\bfseries,
  left=2mm, right=2mm, top=2mm, bottom=2mm
]
\textbf{Query}:\\
Based on the data points [0.07, 0.04, 0.03, -0.01, -0.06, -0.1, -0.11, -0.12, -0.14, -0.17, -0.21, -0.24, -0.29, -0.34, -0.41, -0.5, -0.59, -0.69, -0.76, -0.82, -0.85, -0.87, -0.88, -0.85], predict the likely trend for the next few data points. \\

\textbf{Thought}:\\
I should analyze the trend in the given data to predict the likely direction of future data points. Using the [trend\_classifier] tool with a global analysis will help determine the overall trend.\\

\textbf{Action}:\\
tool: [trend\_classifier], tool\_input: \{"window": null\} \\

\textbf{Observation}: \\
tool: [trend\_classifier], 
\\
output: \{"trend\_results": [\{"channel": 0, "trend": "flat", "analysis\_scope": "global"\}]\} \\

\textbf{Thought}: \\
The [trend\_classifier] indicates a "flat" global trend, which seems inconsistent with the visibly decreasing values in the data. To better capture the recent behavior, I should perform a window-based trend analysis on the latter part of the series where the decline is most pronounced.\\

\textbf{Action}: \\
tool: [trend\_classifier], tool\_input: \{"window": 10\} \\

\textbf{Observation}:\\
tool: [trend\_classifier],\\
output: \{"trend\_results": [\{"channel": 0, "trend": "flat", "segment\_start": 0, "segment\_end": 10, "analysis\_scope": "window=10"\}, \{"channel": 0, "trend": "down", "segment\_start": 10, "segment\_end": 20, "analysis\_scope": "window=10"\}, \{"channel": 0, "trend": "flat", "segment\_start": 20, "segment\_end": 24, "analysis\_scope": "window=10"\}]\}"\}\\

\textbf{Final Answer}:\\
The data shows a recent stabilization after a period of decline. The most likely trend for the next few data points is flat or a very gradual decline, as the sharp downward movement has slowed in the latest segment.
\end{tcolorbox}

\begin{tcolorbox}[
  breakable,
  colback=white,
  colframe=black!50,
  boxrule=0.3mm,
  width=\columnwidth,
  rounded corners,
  title=Sample 5: Pattern Recognition,
  fonttitle=\bfseries,
  left=2mm, right=2mm, top=2mm, bottom=2mm
]
\textbf{Query}:\\
Analyze which statement is correct about the data [39.66, 39.69, 39.91, 40.69, 38.47, 37.5, 36.28, 34.31, 33.12, 33.06, 33.44, 33.72, 33.62, 32.75, 30.44, 29.66, 29.25, 28.5, 28.94, 31.53, 33.0, 35.16, 37.66, 39.62]. \\
A) The data is stable. B) The data shows a clear increasing trend. C) The data mostly decreases with some increase. \\

\textbf{Thought}:\\
To determine which statement is correct, I need to analyze the trend in the data. I will use the [trend\_classifier] tool to classify the global trend of the time series.\\

\textbf{Action}:\\
tool: [trend\_classifier], tool\_input: \{"window": null\} \\

\textbf{Observation}: \\
tool: [trend\_classifier], 
\\
output: \{"trend\_results": [\{"channel": 0, "trend": "flat", "analysis\_scope": "global"\}]\} \\

\textbf{Thought}: \\
The [trend\_classifier] indicates a "flat" global trend. However, visually inspecting the data reveals a clear pattern: it initially increases slightly, then decreases significantly, and finally increases again toward the end. This suggests the global "flat" label may mask underlying dynamics. To better understand, I should examine the data more closely—perhaps by checking for change points or analyzing segments. \\

\textbf{Action}: \\
tool: [change\_point\_detector], tool\_input: \{    "penalty\_or\_n\_cp": "n\_cp=2"\} \\

\textbf{Observation}:\\
tool: [change\_point\_detector],\\
output: change\_point\_results: \{"channel": 0, "change\_point\_indices": [4, 19, 23]\} \\

\textbf{Final Answer}:\\
C) The data mostly decreases with some increase.
\end{tcolorbox}

\begin{tcolorbox}[
  breakable,
  colback=white,
  colframe=black!50,
  boxrule=0.3mm,
  width=\columnwidth,
  rounded corners,
  title=Sample 6: Volatility Evaluation,
  fonttitle=\bfseries,
  left=2mm, right=2mm, top=2mm, bottom=2mm
]
\textbf{Query}:\\
Evaluate the volatility of the data [0.52, 0.51, 0.51, 0.5, 0.49, 0.49, 0.48, 0.48, 0.47, 0.46, 0.46, 0.45, 0.45, 0.44, 0.43, 0.43, 0.42, 0.42, 0.41, 0.41, 0.4, 0.4, 0.39, 0.39]. \\

\textbf{Thought}:\\
I need to compute the rolling volatility (standard deviation of differences) for the provided univariate time series. Since no window size is specified, I will use a reasonable default (e.g., window=5) to capture local variability.\\

\textbf{Action}:\\
tool: [volatility], tool\_input: \{"window": 5\} \\

\textbf{Observation}: \\
tool: [volatility], 
\\
output: \{"window\_size": 5, "volatility\_results": \{"channel\_0": [\{"window\_start": 0, "window\_end": 5, "volatility": 0.00489897948556636\}, \{"window\_start": 1, "window\_end": 6, "volatility": 0.00489897948556636\}, \{"window\_start": 2, "window\_end": 7, "volatility": 0.00489897948556636\}, \{"window\_start": 3, "window\_end": 8, "volatility": 0.00489897948556636\}, \{"window\_start": 4, "window\_end": 9, "volatility": 0.004898979485566351\}, \{"window\_start": 5, "window\_end": 10, "volatility": 0.004898979485566351\} \\

\textbf{Final Answer}:\\
The volatility of the data is consistently low, with a rolling volatility (window size = 5) of approximately 0.0049 across all segments, indicating stable and minimal variability.
\end{tcolorbox}

\textbf{An Application Case Study: Financial Portfolio Risk Monitoring}

To probe usefulness beyond benchmark accuracy, we run TimeART on a portfolio monitoring task that is closer to analyst practice than to multiple-choice question answering. Given six months of daily data for five tickers together with the associated news, the model is asked to produce a risk assessment without any task-specific prompt engineering. Table~\ref{tab:case finance} lists the steps TimeART takes autonomously and their manual counterparts. Analysts judge TimeART's conclusions as substantively equivalent to their own in 92\% of the examined snapshots, where equivalence requires agreement both on the direction of the risk assessment and on the identified drivers, and TimeART needs three minutes per snapshot on average, against two hours for the manual workflow. In one snapshot it surfaced a correlation regime change between two tickers that the manual workflow initially missed. We report this study as qualitative evidence of practical utility, where agreement is judged at the level of conclusions rather than of individual numbers, and we leave a larger blinded study to future work.

\begin{table}[htbp]
\centering
\renewcommand{\arraystretch}{1.2}
\resizebox{\linewidth}{!}{%
\begin{tabular}{@{}lll@{}}
\toprule
\textbf{Analysis step} & \textbf{Tools invoked by TimeART} & \textbf{Manual counterpart} \\
\midrule
Correlation shift detection & \texttt{channel\_correlation}, & correlation matrix \\
 & \texttt{change\_point\_detector} & inspection \\
Return forecasting & \texttt{forecaster} (7-day horizon) & judgement + spreadsheet \\
Risk identification & \texttt{volatility}, \texttt{spike\_detector} & visual chart inspection \\
\bottomrule
\end{tabular}}
\caption{Steps taken autonomously by TimeART on the financial portfolio risk monitoring case, and their manual counterparts.}
\label{tab:case finance}
\end{table}

\section{Failure Case Analysis}
We further analyze the failure cases of TimeART to better understand its current limitations. In analytical tasks, such as forecasting-oriented problems, we observe that the model can generally invoke the appropriate tools. However, some challenging cases still cannot be solved with high precision due to the intrinsic limitations of the tools themselves. For example, forecasting and anomaly detection tools may inevitably introduce prediction errors under noisy, non-stationary, or highly uncertain time-series patterns. Therefore, even when the tool-selection behavior is correct, the final answer may still be affected by the bounded accuracy of the underlying analytical tools.

In reasoning-oriented tasks, failures are more often caused by imperfect tool-use decisions. The model may occasionally select an inappropriate tool, omit a necessary tool invocation, or configure tool parameters incorrectly, leading to flawed intermediate evidence and consequently incorrect reasoning outcomes. We attribute these errors mainly to the limited capacity of the backbone model used in TimeART. Specifically, TimeART is built upon Qwen3 8B, whose capacity may not be sufficient to fully master all tool-use scenarios, especially when the task requires fine-grained parameter control or multi-step reasoning over heterogeneous temporal evidence.

These observations suggest two promising directions for future improvement. First, the analytical reliability of TimeART can be improved by integrating stronger domain tools, such as more advanced forecasting foundation models or more robust anomaly detection modules. Second, the tool-use and reasoning capability can be further enhanced by scaling the backbone LLM, since larger models may better internalize complex tool-selection policies and reduce parameter-configuration errors.

\begin{figure}[!htbp]
  \centering
  \includegraphics[width=\linewidth]{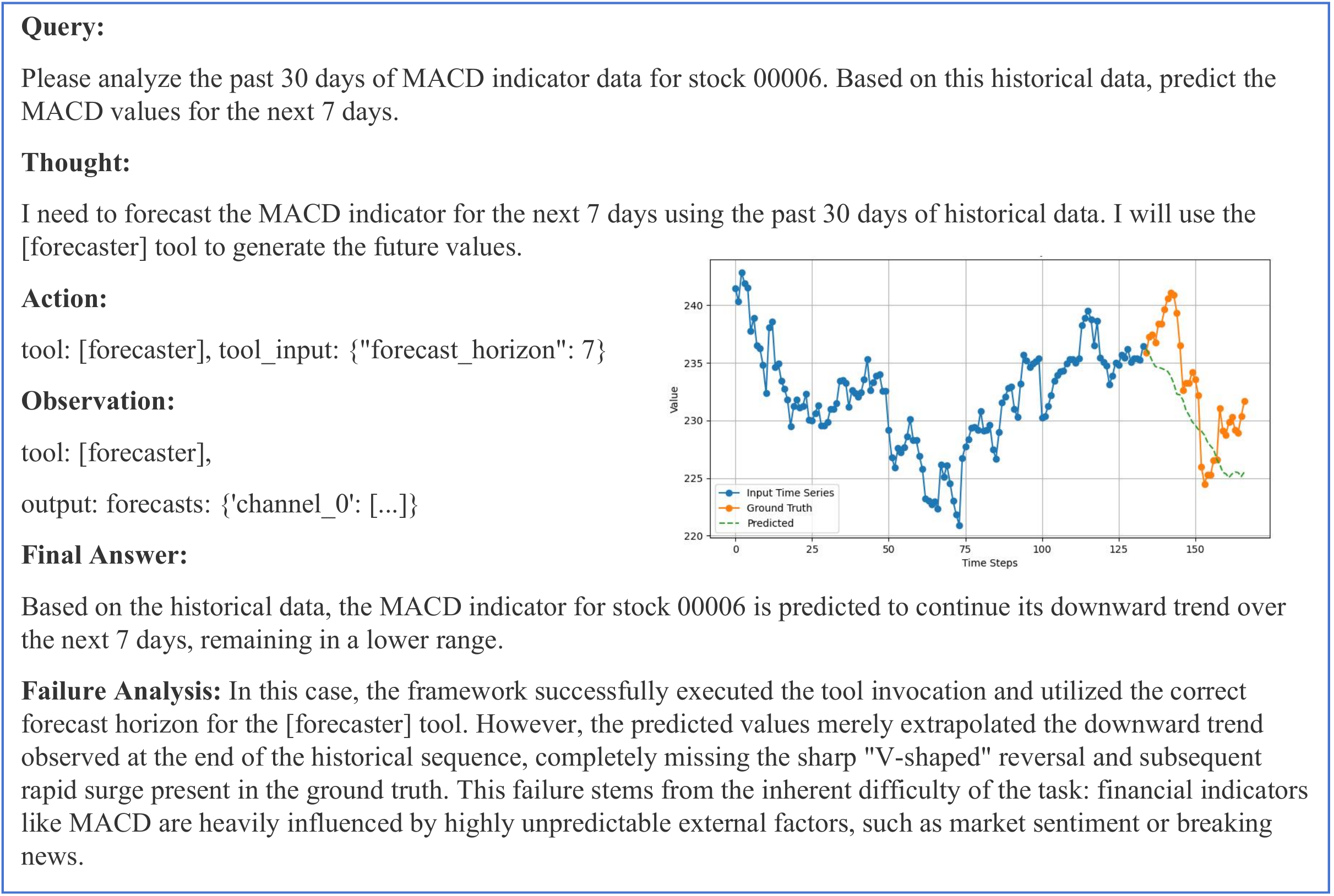}
  \caption{Failure Case a.}
  \label{fig:case a}
\end{figure}

\begin{figure}[!htbp]
  \centering
  \includegraphics[width=\linewidth]{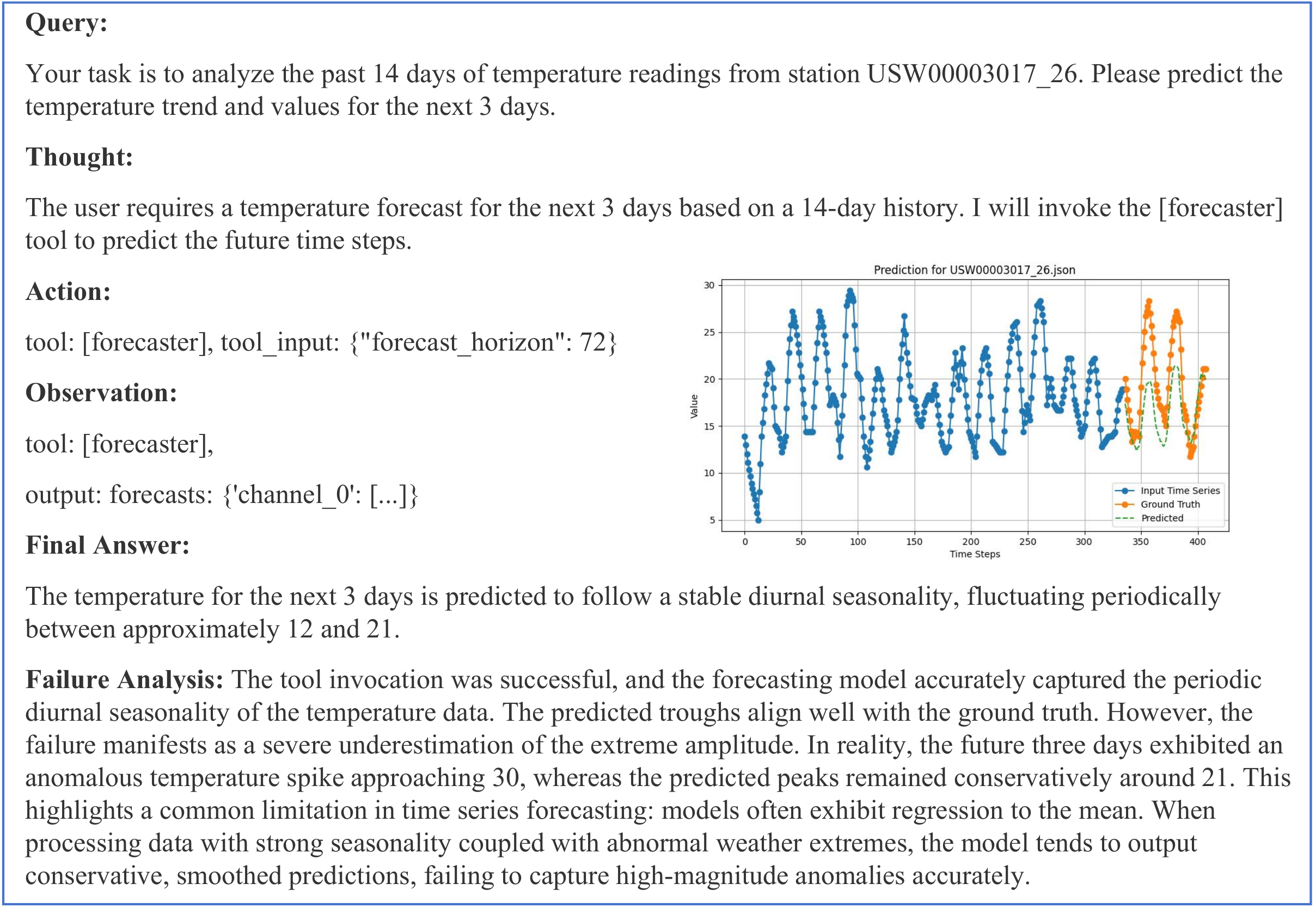}
  \caption{Failure Case b.}
  \label{fig:case b}
\end{figure}

\begin{figure}[!htbp]
  \centering
  \includegraphics[width=\linewidth]{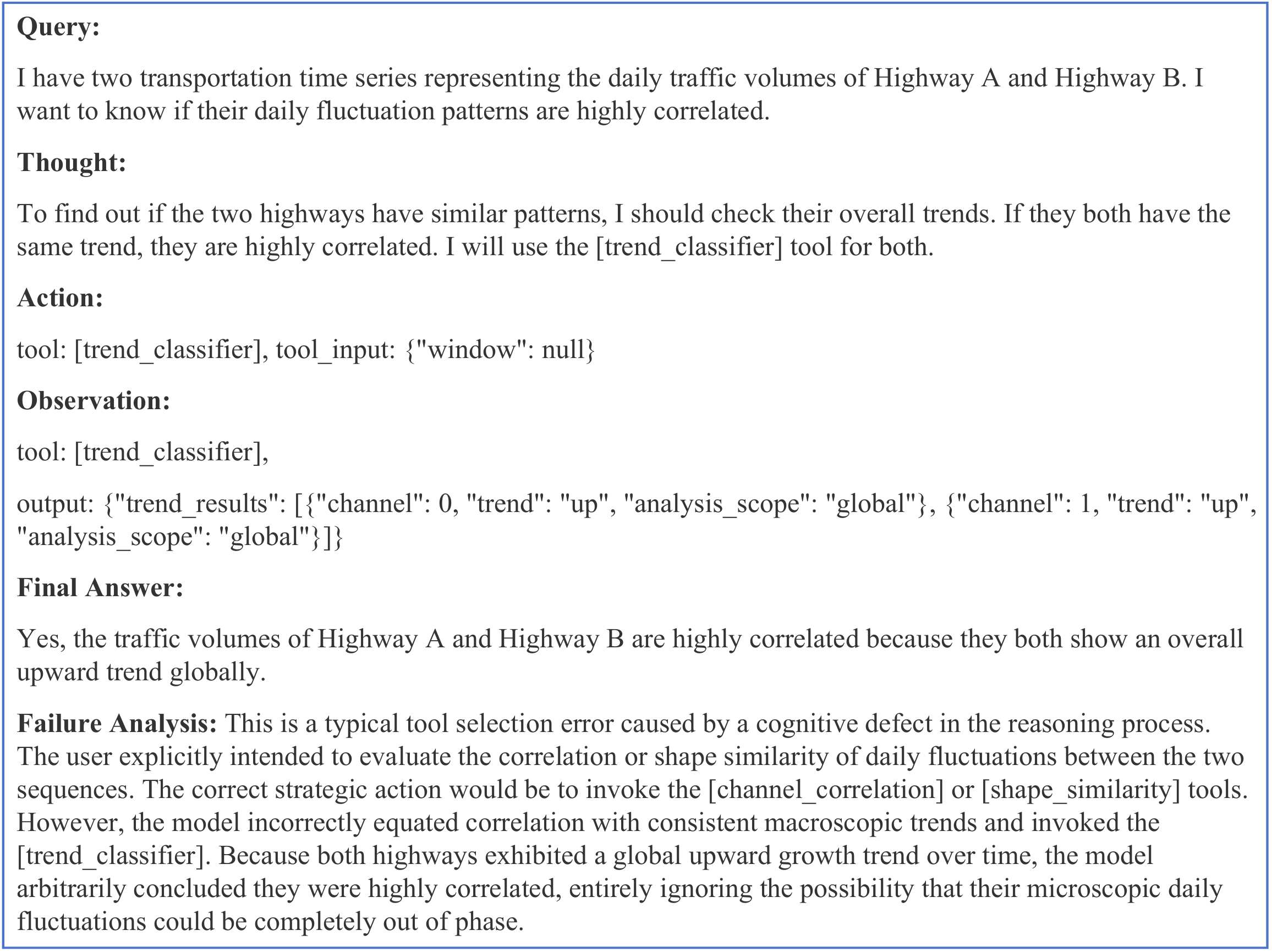}
  \caption{Failure Case c.}
  \label{fig:case c}
\end{figure}

\begin{figure}[!htbp]
  \centering
  \includegraphics[width=\linewidth]{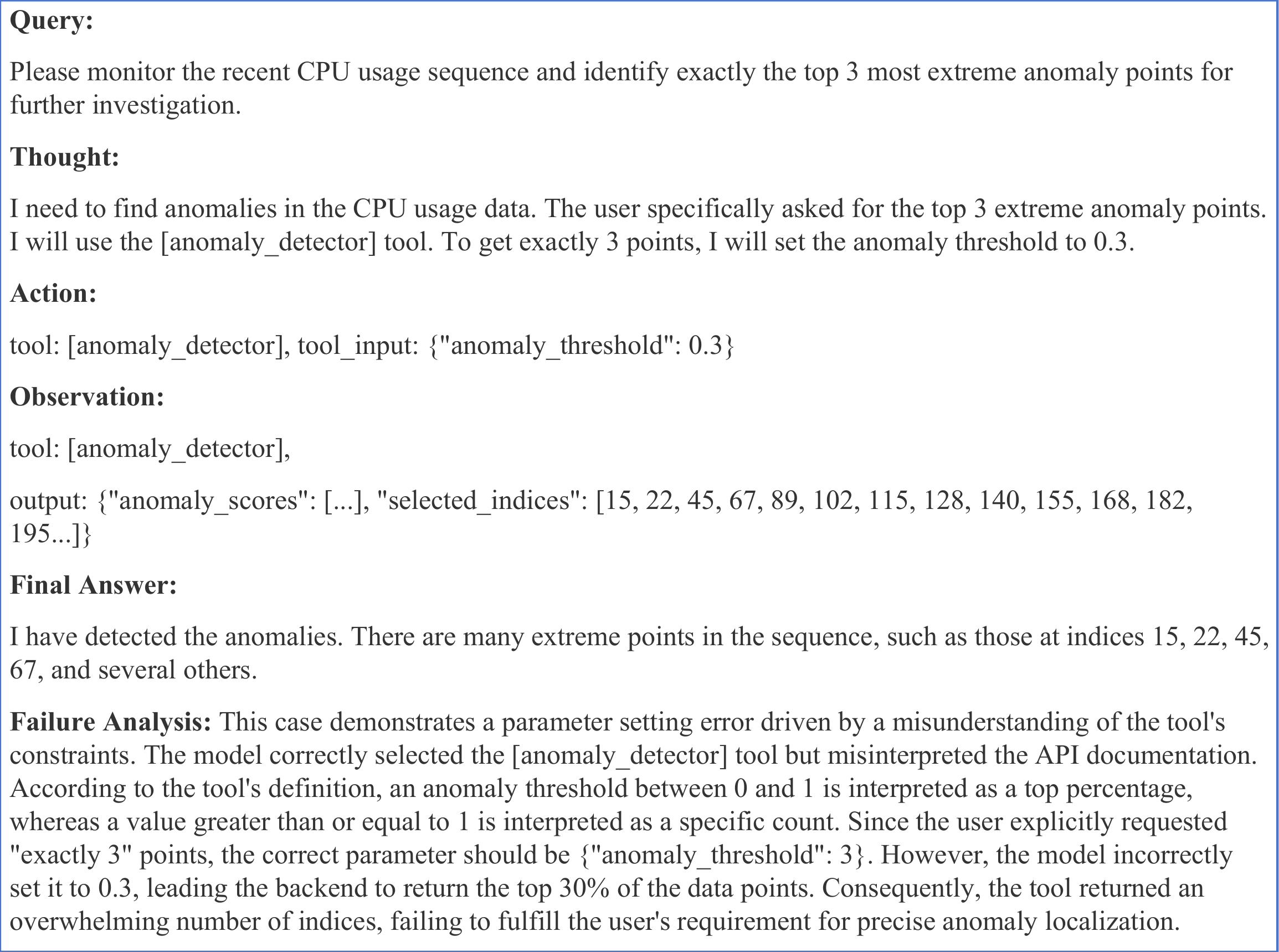}
  \caption{Failure Case d.}
  \label{fig:case d}
\end{figure}

\end{document}